\documentclass[letterpaper]{article} 
\usepackage{aaai2026}  
\usepackage{times}  
\usepackage{helvet}  
\usepackage{courier}  
\usepackage[hyphens]{url}  
\usepackage{graphicx} 
\usepackage{natbib}  
\usepackage{caption} 
\usepackage{algorithm}
\usepackage{algorithmic}
\usepackage{booktabs}
\usepackage{multirow}
\usepackage{amsmath}
\usepackage{amssymb}
\usepackage{bbding}
\usepackage{subcaption}
\usepackage{tcolorbox}
\usepackage{longtable}
\usepackage[printonlyused]{acronym}

\usepackage{newfloat}
\usepackage{listings}
\DeclareCaptionStyle{ruled}{labelfont=normalfont,labelsep=colon,strut=off} 
\floatstyle{ruled}
\newfloat{listing}{tb}{lst}{}
\floatname{listing}{Listing}
\acrodef{LLMs}[LLMs]{large language models}
\acrodef{QA}[QA]{question-answering}
\acrodef{CoT}[CoT]{chain-of-thought}
\acrodef{SOTA}[SOTA]{state-of-the-art}

\title{Large-Scale Diverse Synthesis for Mid-Training}
\author{
    Xuemiao Zhang\textsuperscript{\rm 1,\rm 2}$^{\ast}$, 
    Chengying Tu\textsuperscript{\rm 1,\rm 2}$^{\ast}$, 
    Can Ren\textsuperscript{\rm 1,\rm 2}\thanks{Equal contribution.}, \\
    Rongxiang Weng\textsuperscript{\rm 2}$^{\dagger}$,
    Hongfei Yan\textsuperscript{\rm 1}\thanks{Corresponding author.}, 
    Jingang Wang\textsuperscript{\rm 2}, 
    Xunliang Cai\textsuperscript{\rm 2}
}
\affiliations{

    \textsuperscript{\rm 1} Peking University\quad
    \textsuperscript{\rm 2} Meituan\\
    \{zhangxuemiao, yanhf\}@pku.edu.cn\quad
    \{tuchengying, 2401210098\}@stu.pku.edu.cn\\
    wengrongxiang@gmail.com\quad
    \{wangjingang02, caixunliang\}@meituan.com
}

\usepackage{bibentry}

\begin{document}

\maketitle

\begin{abstract}
The scarcity of high-quality, knowledge-intensive training data hinders the development of \ac{LLMs}, as traditional corpora provide limited information. Previous studies have synthesized and integrated corpora-dependent \ac{QA} data to improve model performance but face challenges in QA data scalability and knowledge diversity, particularly in cross-domain contexts. Furthermore, leveraging our designed discipline and difficulty annotation system, we probe model deficiencies in STEM disciplines and high-difficulty data. To overcome these limitations, we propose a novel diversified pipeline to synthesize BoostQA, a 100B-token large-scale QA dataset. Our synthesis framework:
(1) curates seed data from heterogeneous sources;
(2) utilizes DeepSeek-R1 to implement STEM-focused multi-grade synthesis to boost data diversity and high-difficulty synthesis to mitigate difficulty degradation;  
(3) refines answers via DeepSeek-V3 to improve output quality.
We utilize BoostQA in mid-training, a mid-stage between pre-training and post-training, to optimize domain-specific knowledge acquisition and enhance data quality. 
Our method enables Llama-3 8B, mid-trained on a 40B-token dataset, to achieve an average improvement of \textbf{12.74\%} on MMLU and CMMLU and establish SOTA average performance across 12 benchmarks. BoostQA also demonstrates robust scalability, with performance consistently improving as model size, data volume, and initial FLOPs scale.\footnote{The core code and dataset will be made available at link.}
\end{abstract}


\section{Introduction}

Mid-training has emerged as a pivotal phase in the development of \ac{LLMs}, strategically positioned between pre-training and post-training stages~\citep{wang2025octothinkermidtrainingincentivizesreinforcement}. During this intermediate process, high-quality datasets prove crucial for enhancing the domain-specific capabilities of \ac{LLMs}~\citep{grattafiori2024llama3herdmodels, olmo20252olmo2furious}. Prior research has demonstrated that integrating \ac{QA} data into corpora significantly improves model performance~\citep{parmar2024reusedontretrainrecipe, chen2024towards, wang2025octothinkermidtrainingincentivizesreinforcement}, which is attributable to the high knowledge density and logical organization inherent in the structured QA format. As naturally collected QA data remains insufficient to meet training demands, synthetic methods have become pivotal and effective for QA data generation~\citep{maini2024rephrasing, cheng2024instruction, jiang2025mixcpt, su2024nemotron, akter2025mind, zhou2025megamath}.

\begin{figure}[t]
\centering
\includegraphics[width=\columnwidth]{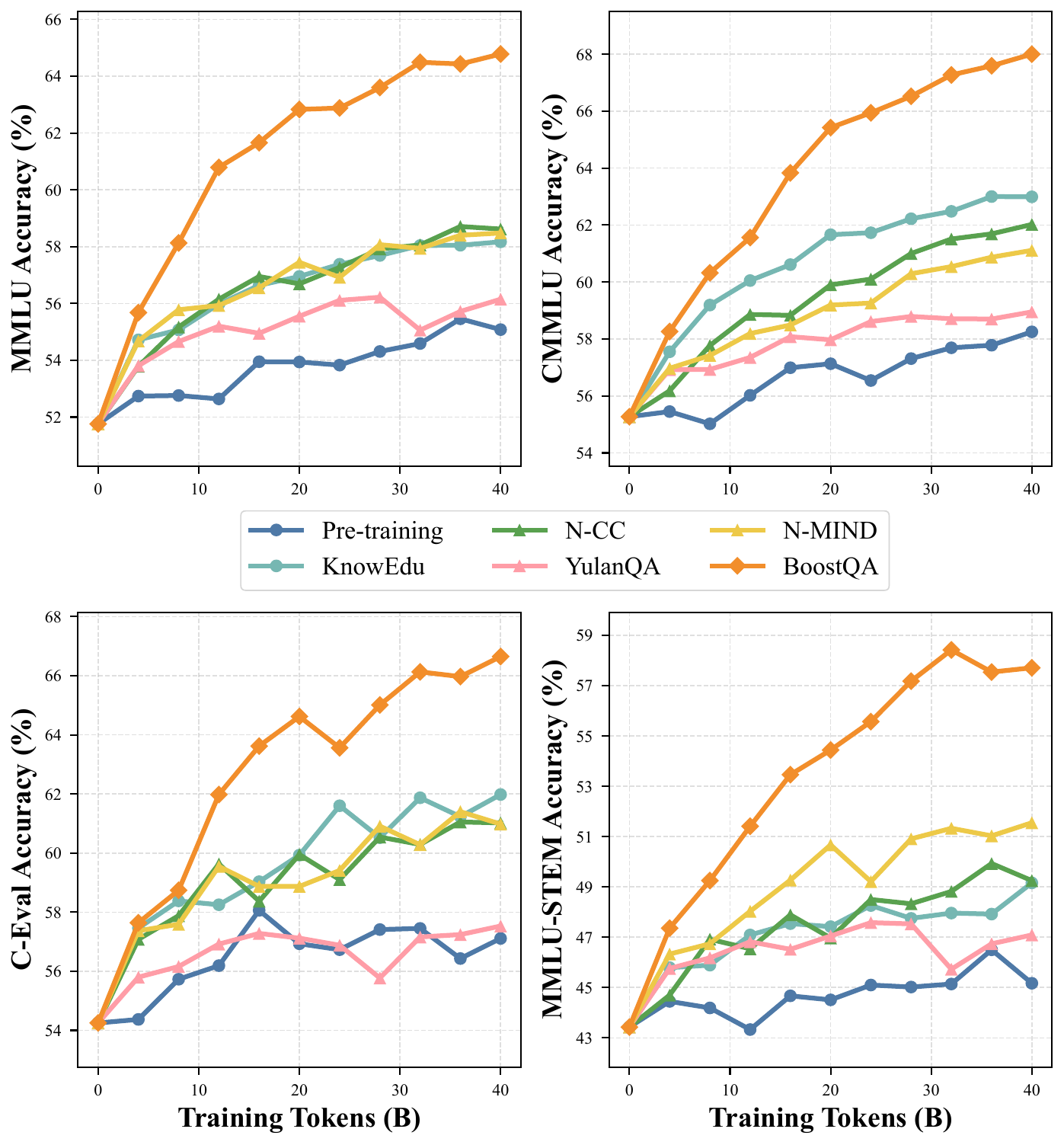} 
\caption{Comparison between BoostQA and baselines.}
\label{fig:mainresult}
\end{figure}

However, existing QA data synthesis methods exhibit critical limitations. Most pipelines operate by leveraging \ac{LLMs} to rephrase or generate QA pairs from corpora, appending them as supplementary material~\citep{maini2024rephrasing, cheng2024instruction, su2024nemotron}. These strategies synthesize QA pairs tightly bound to source passages, which merely extend rather than enrich the original corpora, with text dependency severely constraining QA data reusability. Furthermore, while some open-source synthetic QA datasets are available, most focus on narrow domains, particularly mathematics~\citep{zhou2025megamath, akter2025mind}, and lack both diversity and scale. Consequently, large-scale cross-domain QA datasets remain scarce, limiting model generalization across diverse fields and high-difficulty scenarios that require deep knowledge mastery and reasoning.

Indeed, our probe experiments, as shown in Figure~\ref{fig:test}, demonstrate that the model exhibits differential mastery across QA tasks of varying disciplines and difficulty levels, e.g., achieving 33.78\% accuracy in the highest difficulty level (H5) and 60.74\% in the lowest difficulty level (H1) at the 10T-token checkpoint. This observation motivates a differentiated enhancement strategy, specifically prioritizing the synthesis of STEM disciplines and high-difficulty data.
Building on these insights, we propose a novel pipeline leveraging DeepSeek-R1~\citep{deepseekai2025deepseekr1incentivizingreasoningcapability} to synthesize diverse, STEM-focused, and difficulty-enhanced QA data. We first curate seed data from heterogeneous sources to ensure source diversity. Second, based on our probe results, we design a two-way synthesis pipeline leveraging DeepSeek-R1, comprising a multi-grade synthesizer that generates diverse STEM-focused QA pairs across educational stages and a high-difficulty synthesizer that enhances the synthetic proportion of H4/H5 samples. Finally, we design an answer refinement module powered by DeepSeek-V3~\citep{deepseekai2025deepseekv3technicalreport} to improve output quality. By integrating diversified synthesis factors and high-quality signals from DeepSeek-R1, the pipeline yields BoostQA, a 100B-token large-scale synthetic QA dataset independently of seeds, improving diversity while shifting difficulty distributions upward.

We conduct extensive experiments by mid-training Llama-3 8B~\citep{grattafiori2024llama3herdmodels} on 40B-token data. The results in Figure~\ref{fig:mainresult} demonstrate that BoostQA significantly improves upon the pre-training baseline with an average improvement of \textbf{12.74\%} on MMLU and CMMLU, establishing \ac{SOTA} average performance across 12 benchmarks. Furthermore, performance gains with BoostQA monotonically increase alongside model size, data volume, and initial FLOPs scale, confirming its suitability for large-scale mid-training. In summary, our core contributions are as follows:

\begin{itemize}
    \item We identify deficiencies in model performance related to STEM disciplines and high-difficulty tasks through probe experiments. To address these deficiencies, we propose a novel diversified synthesis pipeline that incorporates diverse seed curation, two-way synthesis focused on STEM and high-difficulty data, and refinement.
    \item We have dedicated substantial resources to developing BoostQA, a plug-and-play, 100B-token large-scale synthetic QA dataset. It specifically targets challenging STEM and high-difficulty questions, advancing the large-scale synthetic data research during mid-training.
    \item Rigorous and comprehensive experimental results confirm that BoostQA achieves SOTA average performance. Specifically, the mid-trained Llama-3 8B boosts average performance by \textbf{12.74\%} on MMLU and CMMLU.
\end{itemize}

\section{Method}
Our synthesis pipeline, illustrated in Figure~\ref{fig:method}, is developed by addressing fundamental challenges in QA synthesis. We identify model deficiencies through probe experiments to guide the synthesis process while maintaining rigorous quality control. The seed data is sourced from QA pairs~\citep{li2016datasetneuralrecurrentsequence, amini-etal-2019-mathqa}, books~\citep{gao2020pile800gbdatasetdiverse}, and high-quality web pages~\citep{chang2024redstonecuratinggeneralcode, zhou2025megamath} to ensure diversity of sources. To avoid contamination, we follow previous work~\citep{shao2024deepseekmathpushinglimitsmathematical} to conduct exact 10-gram matching and embedding-based similarity to filter out seeds containing questions or answers to any of our evaluation benchmarks.

\subsection{Model Probes to Define Synthesis Direction}

Traditional synthesis methods often produce undifferentiated data, failing to address specific model deficiencies~\citep{chen2024towards, su2024nemotron}. To overcome this limitation, we establish a dual annotation system that combines standardized discipline classification with human-calibrated difficulty scoring. We fine-tune Qwen2.5-7B-Instruct~\citep{qwen2025qwen25technicalreport} distilled from DeepSeek-R1 to obtain a discipline classifier capable of classifying content into 62 distinct disciplines. In parallel, the difficulty scorer employs human performance metrics, defined as pass rates under standardized one-hour testing conditions with QS Top 100 university students majoring in relevant disciplines, to calibrate five difficulty tiers (H1-H5), as displayed in Figure~\ref{fig:diffdef}. Further details can be found in Appendix~\ref{sec:appendix_annotation}.

\begin{figure}[h]
\centering
\includegraphics[width=\columnwidth]{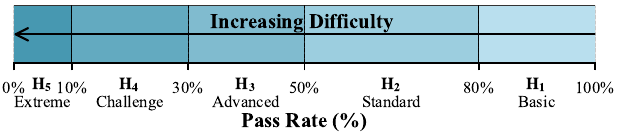}
\caption{Difficulty levels definition.}
\label{fig:diffdef}
\end{figure}

\begin{figure}[t]
\centering
\begin{subfigure}{0.48\linewidth}
    \centering
    \includegraphics[width=\linewidth, height=5cm, keepaspectratio]{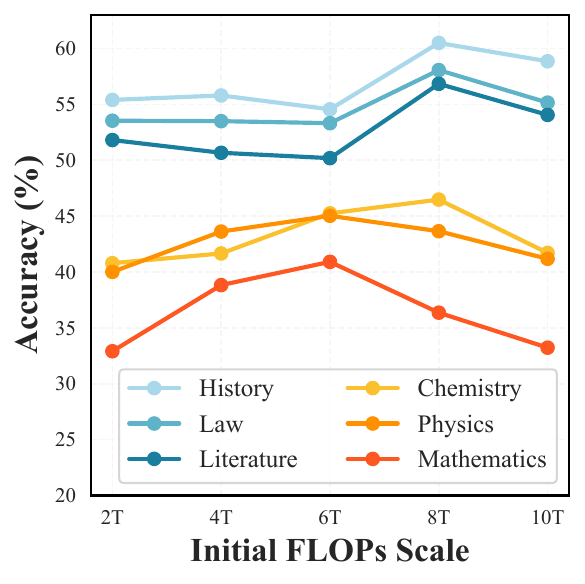}
    \caption{Discipline probe result.}
    \label{fig:subtest}
\end{subfigure}
\begin{subfigure}{0.48\linewidth}
    \centering
    \includegraphics[width=\linewidth, height=5cm, keepaspectratio]{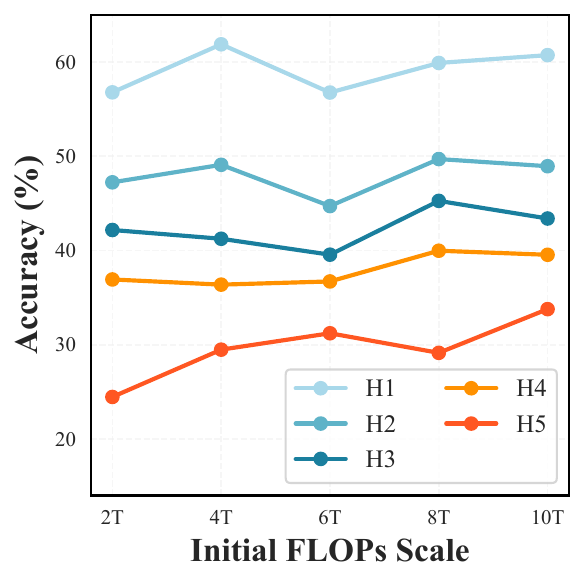}
    \caption{Difficulty probe result.}
    \label{fig:difftest}
\end{subfigure}

\caption{The discipline and difficulty probe experimental results, with six typical disciplines shown.}
\label{fig:test}
\end{figure}

This dual annotation enables probe experiments to evaluate model mastery of QA seeds across different disciplines and difficulty levels at various initial FLOPs from the 2T-token checkpoint to the 10T-token checkpoint (setup detailed in \textsection~\ref{sec:train_details}). Following SliCK~\citep{gekhman-etal-2024-fine} settings, we evaluate model performance over 10 repeated trials, each utilizing a distinct 5-shot prompt in greedy generation mode. The aggregate accuracy across all trials serves as the metric to assess model mastery of the given data. Based on the results in Figure~\ref{fig:test}, we summarize \textbf{two key findings}:
For the discipline probes, all models exhibit higher accuracy in humanities and lower accuracy in STEM disciplines;
For the difficulty probes, there is a monotonic degradation in accuracy from H1 to H5 levels. 
Therefore, these findings inspire us to prioritize the synthesis of STEM and high-difficulty (H4/H5) data, consistent with prior research~\citep{parmar2024reusedontretrainrecipe, chen2024towards}.

\subsection{Enhance Diversity via Educational Role}
Current synthetic QA datasets often lack diversity and scale, primarily due to the singularity of their synthesis methods, which typically involve only rephrasing or text questioning by \ac{LLMs}~\citep{maini2024rephrasing, cheng2024instruction, su2024nemotron}. Previous studies have made preliminary explorations into multi-role prompts~\citep{ge2025scalingsyntheticdatacreation}. Building upon this, we employ DeepSeek-R1 and design educational role paradigms for multi-grade synthesis. These paradigms include roles such as high school, college, and graduate-level teachers, modulating linguistic complexity and conceptual depth through tier-specific prompting strategies to enhance the scale of data diffusion.

As illustrated in Figure~\ref{fig:method}, the settings for multi-grade synthesis are governed by three key components. The rule enforcer manages the structural formats for two types of questions through specific constraints. For multiple-choice questions, it mandates distractor-optimized formats with four options and clear correct answers. For essay questions, it requires self-contained questions that necessitate explanatory responses, thus preventing ambiguous formulations. Configurable input settings allow adjustments to seed data formats and output volume parameters, such as $n=10$, with enhanced sampling proportion of STEM seeds. The corresponding prompt is provided in Appendix~\ref{sec:appendix_prompts_synthesis}. Strategic combinations of these components enhance diversity.

\subsection{Enhance to Mitigate Difficulty Degradation}

\begin{figure}[t]
\centering
\includegraphics[width=0.9\columnwidth]{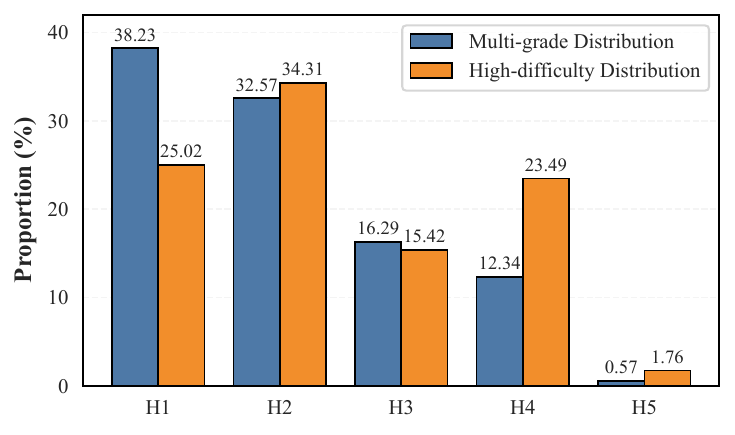} 
\caption{Difficulty distribution of multi-grade synthetic data and high-difficulty synthetic data.}
\label{fig:diffcmp}
\end{figure}

Post-validation employs DeepSeek-V3 to relabel the \textit{actual} educational stage of a sampled multi-grade synthetic subset, shown in Table~\ref{tab:findings_stage} in Appendix~\ref{sec:appendix_findings_edu}. This reveals a systematic downward degradation in educational stage alignment within synthetic QA data. To characterize this phenomenon, we employ the difficulty scorer to quantify the distribution profile, which exhibits a significant skew towards lower difficulty levels, as visualized in Figure~\ref{fig:diffcmp}. This distributional bias indicates inherent limitations in generating difficult content through standard synthesis methods.

To address this deficiency, we reconfigure the synthesis architecture to prioritize high-difficulty (H4/H5 tiers) seeds based on enhanced STEM sampling. As shown in Figure~\ref{fig:method}, the role assigner is restricted to the graduate level, as detailed in Appendix~\ref{sec:appendix_prompts_synthesis}, while a novel difficulty booster module is integrated to establish a dedicated high-difficulty synthesizer. The difficulty booster improves the sampling weight of high-difficulty seeds and incorporates the architecture of the difficulty scorer, explicitly requiring QA pairs to high-difficulty H4/H5 tiers during data generation. This architectural refinement specifically targets amplification of educationally challenging QA pairs through controlled expertise simulation and difficulty augmentation. Empirical evaluation in Figure~\ref{fig:diffcmp} demonstrates substantial improvement in synthetic QA difficulty, showing a $1.96\times$ increase in high-difficulty H4/H5 synthetic proportion, rising from 12.91\% to 25.25\% compared to the multi-grade synthesizer.

\subsection{Answer Refinement and Blend with Corpora \label{sec:method_blend_corpora}}
\begin{figure*}[t]
  \centering
  \includegraphics[width=\textwidth]{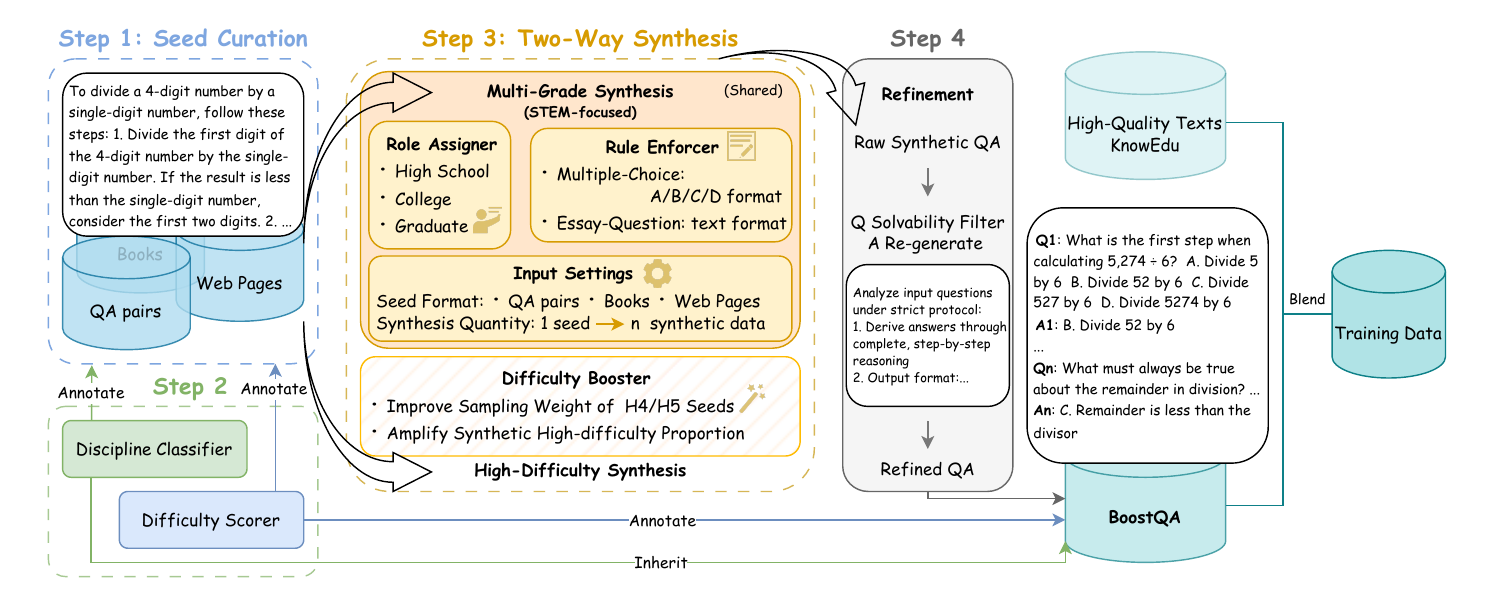}
  \caption{The pipeline for synthesizing BoostQA. In Step 1, diverse high-quality seed data is collected, comprising QA pairs, books, and quality web pages. In Step 2, we annotate the seed data with discipline and difficulty labels to conduct probe experiments, identify model deficiencies, and determine synthesis direction. In Step 3, we implement a two-way synthesis strategy: STEM-focused multi-grade synthesis uses simulated educational roles to enhance diversity, while high-difficulty synthesis incorporates an additional difficulty booster to increase the proportion of high-difficulty synthetic data. In Step 4, answer refinement is applied to synthesized QA pairs. Finally, the generated BoostQA is blended with the high-quality general corpora, KnowEdu, for mid-training.}
  \label{fig:method}
\end{figure*}

Based on previous synthesis work~\citep{zhou2025megamath}, we have incorporated answer refinement into our synthesis pipeline. This refinement module employs a rigorous verification process for synthesized QA pairs, as illustrated in Figure~\ref{fig:method}. Specifically, for each QA pair, we first utilize DeepSeek-V3 to assess the solvability of the question and filter out unsolvable questions due to missing critical information, erroneous conditions, or similar issues. For solvable questions, a re-analysis is performed through stepwise reasoning to generate a refined answer. We additionally perform the same decontamination for synthetic data as for seeds.

The complete pipeline is depicted in Figure~\ref{fig:method}. To ensure distributional alignment with standard pre-training corpora while maintaining data quality~\citep{parmar2024reusedontretrainrecipe}, we construct knowledge-intensive corpora named KnowEdu through a dual-criteria selection protocol and blend it with BoostQA to train the model. The pre-training dataset undergoes rigorous quality stratification using the QuRater model~\citep{wettig2024qurating} to quantify knowledge density, alongside the educational classifier from FineWeb-Edu~\citep{penedo2024fineweb} to assess educational utility. Texts rated highly in both knowledge density and educational utility are retained to form KnowEdu. When blended with synthetic QA data, KnowEdu serves critical functions: anchoring the hybrid distribution to established pre-training data manifolds to mitigate catastrophic forgetting, and providing complementary discursive context absent in QA structures.


\section{Experiments and Results}
\subsection{Experimental Setup}

\paragraph{Training Details.\label{sec:train_details}}
We conduct experiments on Llama-3 8B, which is pre-trained on 10T tokens. The pre-training dataset $\mathcal{D}_{\text{pt}}$ is sourced from diverse domains, mirroring the composition of the Matrix dataset~\citep{zhang2024mapneohighlycapabletransparent}. 
Following the data blend strategy of~\citet{parmar2024reusedontretrainrecipe}, we construct the mid-training dataset $\mathcal{D}_{\text{mt}}$ as a 1:1 blend of high-quality general text $\mathcal{D}_{\text{text}}$ and QA data $\mathcal{D}_{\text{qa}}$. We utilize KnowEdu as $\mathcal{D}_{\text{text}}$, detailed in \textsection~\ref{sec:method_blend_corpora}, and $\mathcal{D}_{\text{qa}}$ adopts the naive format \textit{\{question\}\textbackslash n\{answer\}}, without \ac{CoT} reasoning. 
We fine-tune a synthesis model distilled from DeepSeek-R1 and an answer refinement model from DeepSeek-V3, with model inference executed on multiple H20 GPUs. 
Our main experiments commence mid-training from the 2T-token checkpoint using 40B tokens of $\mathcal{D}_{\text{mt}}$, implemented via the Megatron framework~\citep{shoeybi2020megatronlmtrainingmultibillionparameter}. We utilize the Adam optimizer, with a linearly decaying learning rate schedule initialized at $1.9\times 10^{-4}$ and terminating at $1.9\times 10^{-5}$, as detailed in Appendix~\ref{sec:appendix_train_details}.

Further ablation studies systematically examine scalability in the following aspects: model size, data volume, and initial FLOPs.
For model size scalability, we evaluate Llama-3 models with 1.7B, 8B, and 16B parameters under identical 40B-token $\mathcal{D}_{\text{mt}}$ configurations. 
For data volume scalability, we expand $\mathcal{D}_{\text{mt}}$ from 40B to 190B tokens for Llama-3 8B. 
For initial FLOPs scalability, we compare checkpoints initialized at the 2T-token checkpoint versus the 10T-token checkpoint for Llama-3 8B.
All ablations preserve the learning rate decay relative to their respective pre-training checkpoints.

For math domain experiments, we retain $\mathcal{D}_{\text{text}}$ as KnowEdu while modulating $\mathcal{D}_{\text{qa}}$ across three variants: BoostQA$_{\text{MegaMath}}$, synthesized from high-quality web pages MegaMath-Web-Pro~\citep{zhou2025megamath} seeds without \ac{CoT} reasoning; BoostQA$_{\text{Math}}$, synthesized from mathematical seeds without \ac{CoT} reasoning; BoostQA$_{\text{MathCoT}}$, augmented from BoostQA$_{\text{Math}}$ with \ac{CoT} reasoning in the \textit{\{question\}\textbackslash n\{CoT\}\textbackslash n\{answer\}} format.
\paragraph{Evaluation.}
Our comprehensive evaluation framework leverages lm-evaluation-harness~\citep{gao2021framework} to assess model capabilities across multiple dimensions. 
Knowledge-intensive evaluation includes MMLU~\citep{hendrycks2021measuring}, CMMLU~\citep{li-etal-2024-cmmlu}, C-Eval~\citep{huang2023ceval}, MMLU-Pro~\citep{wang2024mmlupro}, and MMLU-STEM. Mathematical reasoning evaluation encompasses GSM8K~\citep{cobbe2021training} and MATH~\citep{hendrycks2021measuringmath}. Commonsense reasoning evaluation comprises WinoGrande~\citep{sakaguchi2021winogrande}, HellaSwag~\citep{zellers-etal-2019-hellaswag}, and ARC-C~\citep{clark2018think}. Complex reasoning evaluation includes BIG-Bench~\citep{suzgun-etal-2023-challenging} and DROP~\citep{dua-etal-2019-drop}. 
\begin{table*}
    \centering
    \setlength{\tabcolsep}{1mm}
    \begin{tabular}{lcccccccccccccc}
        \toprule
        \textbf{Dataset} & \textbf{MMLU} & \textbf{CMMLU} & \textbf{C-Eval} & \textbf{M-Pro} & \textbf{STEM} & \textbf{GSM8K} & \textbf{MATH} & \textbf{W.G.} & \textbf{H.S.} & \textbf{ARC-C} & \textbf{BBH} & \textbf{DROP} & \textbf{AVG.}\\
        \midrule
        Pre-training & 55.08 & 52.23 & 57.11 & 24.32 & 45.17 & 33.95 & 6.50 & 51.50 & \underline{43.00} & 70.50 & 35.79 & 42.31 & 43.12 \\
        FineWeb-Edu & 56.23 & 58.88 & 56.80 & 25.46 & 47.78 & 31.49 & 2.50 & 53.50 & 35.00 & 69.60 & 34.38 & 39.44 & 42.59 \\
        KnowEdu & 58.17 & \underline{62.99} & \underline{61.98} & 25.64 & 49.16 & 32.56 & 8.00 & 54.50 & 36.00 & 71.50 & 35.12 & 41.07 & 44.72 \\
        \midrule
        N-CC  & \underline{58.62} & 62.02 & 59.02 & 27.68 & 49.25 & 29.33 & 7.00 & 55.00 & 41.50 & \underline{73.00} & 34.86 & 41.63 & 44.91 \\
        YulanQA & 56.15 & 58.95 & 57.53 & 24.89 & 47.09 & 40.48 & 9.00 & 53.00 & \underline{43.00} & \underline{73.00} & 35.75 & 42.18 & 44.25 \\
        N-MIND & 58.48 & 61.11 & 60.98 & \underline{30.32} & \underline{51.55} & 47.42 & 13.50 & 52.00 & 40.50 & \underline{73.00} & \underline{37.69} & \underline{46.33} & 47.74 \\
        MegaMathQA & 55.98 & 59.04 & 58.91 & 26.86 & 48.59 & 44.65 & 6.50 & \underline{55.50} & 41.50 & 68.50 & 35.64 & 43.31 & 45.42 \\
        JiuZhang3.0 & 56.55 & 60.30 & 59.52 & 27.43 & 48.68 & \textbf{56.27} & \textbf{23.00} & 55.00 & 36.50 & 71.50 & 36.33 & 45.05 & \underline{48.01} \\
        \midrule
        BoostQA & \textbf{64.78} & \textbf{68.00} & \textbf{66.65} & \textbf{30.61} & \textbf{57.71} & \underline{47.73} & \underline{14.00} & \textbf{60.50} & \textbf{44.00} & \textbf{80.00} & \textbf{38.52} & \textbf{47.41} & \textbf{51.66} \\
        \bottomrule
    \end{tabular}
    \caption{Comparison across 12 benchmarks. The best and second best are in bold and underlined, respectively. Abbreviations: M-Pro = MMLU-Pro, STEM = MMLU-STEM, W.G. = WinoGrande, H.S. = HellaSwag, BBH = Big-Bench.}
    \label{tab:mainresult}
\end{table*}
\paragraph{Baselines.}
Baselines are categorized into two paradigms. The first evaluates the 40B-token general corpora, comprising: the pre-training dataset $\mathcal{D}_{\text{pt}}$, utilized for further mid-training; FineWeb-Edu~\citep{penedo2024fineweb}, which consists of web-crawled, high-quality educational texts; and KnowEdu, our curated high-quality knowledge-rich and educational data. 
The second paradigm assesses the QA blend following a 1:1 mixing ratio between KnowEdu and QA datasets. For general-domain baselines: Nemotron-CC (N-CC)~\citep{su2024nemotron}, with 9:1 document-to-QA ratio, and YulanQA, a collected QA subset from the continual pre-training dataset of~\citet{chen2024towards}. 
For mathematics baselines: Nemotron-MIND (N-MIND)~\citep{akter2025mind}, a synthetic dataset of math dialogues; MegaMathQA, QA subset from MegaMath-Synthetic~\citep{zhou2025megamath}; and JiuZhang3.0~\citep{zhou2024jiuzhang}, structured math QA with \ac{CoT}. 
Information about the QA datasets can be found in Table~\ref{tab:appendix_datasets} in Appendix~\ref{sec:appendix_datasets}. 


\subsection{Main Results}

Figure~\ref{fig:mainresult} and Table~\ref{tab:mainresult} present our main results, with MMLU subsets results detailed in Appendix~\ref{appendix_findings_mmlusubsets}. We find that:

\paragraph{BoostQA can significantly boost the performance of LLMs.} Compared to the pre-training baseline, BoostQA demonstrates average improvements of \textbf{12.74\%} on MMLU and CMMLU, and an average gain of 8.54\% across 12 benchmarks. Even against high-quality corpora, FineWeb-Edu and KnowEdu, BoostQA offers improvements of 9.07\% and 6.94\% on average.

\paragraph{BoostQA surpasses QA baselines and establishes SOTA performance.} On MMLU and CMMLU, BoostQA achieves an average improvement of 6.07\% over the strongest QA baseline, N-CC, and an improvement of 3.65\% over JiuZhang3.0 across 12 diverse benchmarks. In knowledge-intensive evaluation, BoostQA significantly outperforms all baselines, demonstrating exceptional cross-domain understanding. Furthermore, it universally leads in commonsense and complex reasoning benchmarks. This comprehensive advantage stems from our novel synthesis pipeline that systematically integrates educational diversity and a difficulty booster, in conjunction with knowledge distilled from DeepSeek-R1. 

\begin{figure}[t]
\centering
\includegraphics[width=0.9\columnwidth]{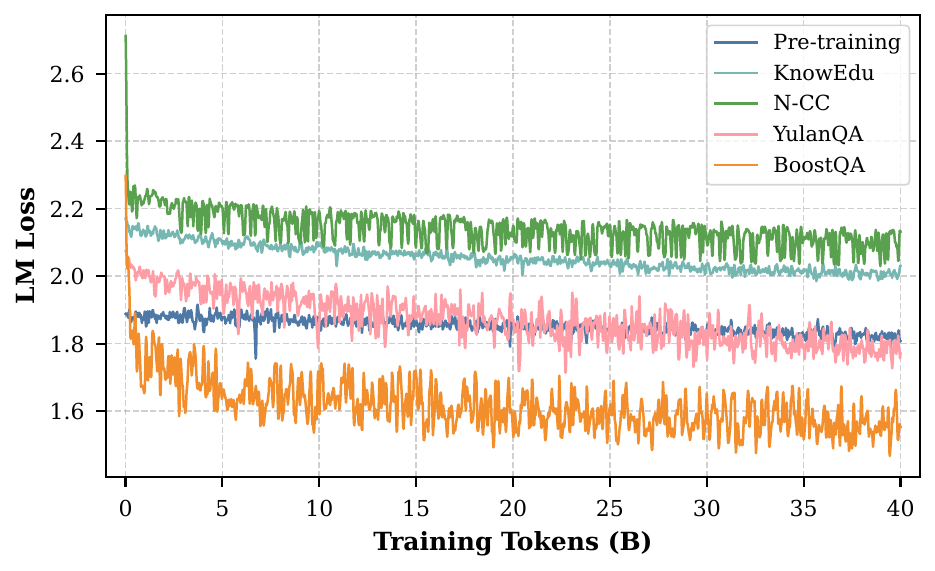} 
\caption{Smoothed LM loss over training tokens.}
\label{fig:loss}
\end{figure}

\paragraph{BoostQA exhibits strong scalability and stability.} The longitudinal analysis, illustrated in Figure~\ref{fig:mainresult}, reveals that as training progresses, BoostQA consistently maintains performance leadership across benchmarks. Notably, its upward trajectory continues even at the 40B-token threshold, indicating substantial unexploited potential. This training stability and continued improvement contrast sharply with other baselines, which exhibit convergence, highlighting BoostQA's unique capacity for sustainable capability growth through large-scale data expansion. 


\paragraph{The training loss results illustrated in Figure~\ref{fig:loss} demonstrate that BoostQA achieves the lowest final loss.} 
This observation aligns with BoostQA's superior performance and is consistent with the loss-performance correlation reported by~\citet{du2024understanding}. Notably, all QA blends, including N-CC, YulanQA, and BoostQA, exhibit heightened loss volatility. This may arise from structural conflicts between knowledge-intensive QA pairs and descriptive general text.

\begin{figure*}[t]
\centering
\includegraphics[width=0.9\textwidth]{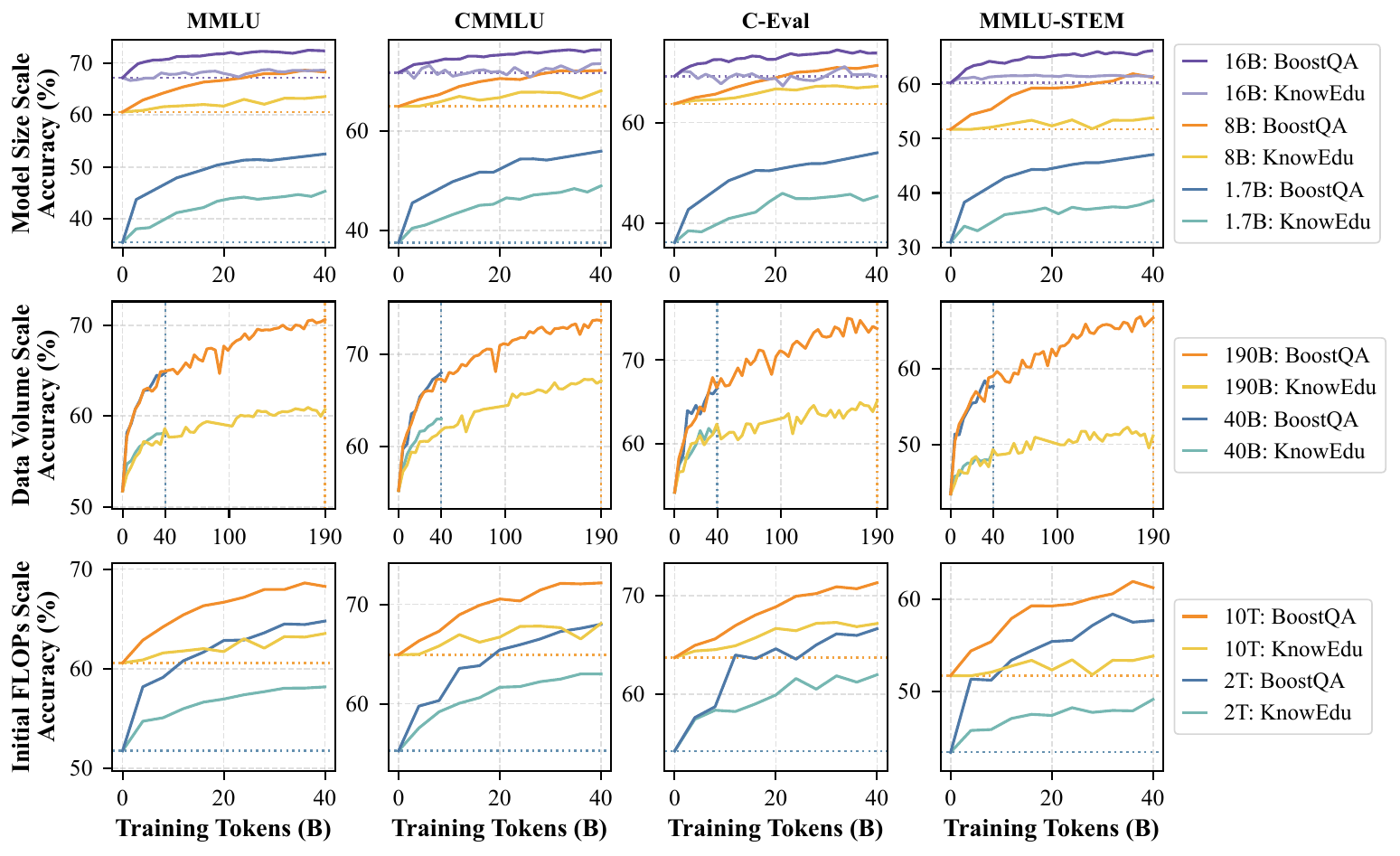}
\caption{Multi-dimensional scalability of BoostQA, detailed values shown in Table~\ref{tab:appendix_scaling} in Appendix~\ref{sec:appendix_scaling}. For model size scalability, the initial checkpoints for the models with 1.7B, 8B, and 16B parameters are 4T, 10T, and 10T tokens, respectively. For data volume scalability, both initial checkpoints are 2T tokens.}
\label{fig:scale}
\end{figure*}

\subsection{Scalability Analysis \label{sec:scaling}}
To verify the robustness of BoostQA, we conduct scalability experiments across three dimensions, as shown in Figure~\ref{fig:scale}.

\paragraph{At various model sizes, the performance gap between BoostQA and the baseline progressively widens as training advances.} This suggests that BoostQA provides complementary learning signals that accumulate over time, rather than offering temporary benefits. BoostQA demonstrates measurable performance improvements over KnowEdu for all model sizes. These enhancements are consistent across all evaluated benchmarks, with the magnitude of improvement positively correlating with increased exposure to training tokens.

\paragraph{At different data volumes, BoostQA maintains significant effectiveness when extending to 190B tokens.} The model's performance continues to rise at both the 40B and 190B tokens, demonstrating the strong scalability of BoostQA. Upon completing full-scale training, the model trained with BoostQA on 190B tokens surpasses the one trained on 40B tokens, improving by an average of 5.76\% on MMLU and CMMLU.

\paragraph{At different initial FLOPs, BoostQA can effectively augment training performance.} Varying initial FLOPs reflect the different capabilities of the model at the start of mid-training. Introducing BoostQA at either the 2T-token (early-stage) or 10T-token (final-stage) checkpoint yields stable performance gains compared to KnowEdu. Notably, models trained with BoostQA from the 2T-token checkpoint ultimately surpass the performance of models trained with KnowEdu from the 10T-token checkpoint on MMLU.

\subsection{Mathematical Results \label{sec:exp_math}}

\begin{table}[t]
    \centering
    \setlength{\tabcolsep}{1mm}
    \small
    \begin{tabular}{lcccccccc}
        \toprule
        \textbf{Dataset} & \textbf{CoT} & \textbf{G.} & \textbf{M.} & \textbf{Elem.} & \textbf{High.} & \textbf{Coll.} & \textbf{AVG.} \\
        \midrule
        Pre-training & - & 33.95 & 6.50 & 36.50 & 30.50 & 25.00 & 26.49 \\
        FineWeb-Edu & - & 31.49 & 2.50 & 38.00 & 31.00 & 30.00 & 26.60 \\
        KnowEdu & - & 32.56 & 8.00 & 37.50 & 27.50 &31.00 & 27.31 \\
        \midrule
        N-MIND & \CheckmarkBold & 47.42 & 13.50 & 43.50 & 29.50 & 37.00 & 34.18 \\
        MegaMathQA & \CheckmarkBold & 44.65 & 6.50 & 47.00 & 31.50 & 35.00 & 32.93 \\
        JiuZhang3.0 & \CheckmarkBold & \underline{56.27} & \textbf{23.00} & 42.50 & 30.50 & 31.00 & 36.65\\
        \midrule
        BoostQA & \XSolidBrush & 47.73 & 14.00 & 53.00 & 44.50 & 38.00 & 39.45 \\ 
        BoostQA$_\text{MM}$ & \XSolidBrush & 42.26 & 13.50 & \textbf{60.50} & \textbf{49.50} & \textbf{48.00} & \underline{42.75} \\
        BoostQA$_\text{Math}$ & \XSolidBrush & 42.96 & 13.50 & \underline{57.00} & \underline{46.50} & \underline{44.00} & 40.79 \\
        BoostQA$_\text{MC}$ & \CheckmarkBold & \textbf{65.67} & \underline{22.50} & 54.50 & 46.00 & 43.00 & \textbf{46.33} \\
        \bottomrule
    \end{tabular}
    \caption{Comparison of mathematical performance. The best and second best are in bold and underlined, respectively. Abbreviations: BoostQA$_\text{MM}$ = BoostQA$_\text{MegaMath}$, BoostQA$_\text{MC}$ = BoostQA$_\text{MathCoT}$, G. = GSM8K, M. = MATH, MMLU mathematics subsets: Elem. = elementary, High. = high school, Coll. = college.}
    \label{tab:mathresult}
\end{table}

\paragraph{BoostQA$_\text{MathCoT}$ with \ac{CoT} achieves SOTA average performance across mathematical benchmarks.} As shown in Table~\ref{tab:mathresult}, BoostQA$_\text{MegaMath}$ without \ac{CoT} attains optimal accuracy across all mathematical subsets of MMLU. BoostQA$_\text{MathCoT}$ delivers peak performance on GSM8K while achieving near-parity to JiuZhang3.0 with \ac{CoT} on the challenging MATH benchmark. BoostQA$_\text{Math}$ maintains competitive performance across all mathematical subsets of MMLU, despite weaknesses in GSM8K and MATH.

\paragraph{The comparative advantage of BoostQA$_\text{MathCoT}$ on mathematical benchmarks underscores the essential role of \ac{CoT}.} As indicated in Table~\ref{tab:mathresult}, BoostQA$_\text{MathCoT}$ outperforms other BoostQA variants without \ac{CoT} on GSM8K and MATH. When compared with mathematical baselines with \ac{CoT}, BoostQA$_\text{MathCoT}$ not only maintains comparable performance on GSM8K and MATH, but also exhibits an advantage in the mathematical subsets of MMLU.

\paragraph{BoostQA$_\text{MegaMath}$ demonstrates advantages over MegaMathQA, despite being derived from the same seed data.} As shown in Table~\ref{tab:mathresult}, BoostQA$_\text{MegaMath}$ outperforms MegaMathQA by an average of 9.82\%. This comparison underscores the effectiveness of BoostQA's distinctive knowledge transformation mechanics. While MegaMathQA relies on conventional extraction and refinement techniques, our synthesis pipeline fundamentally diffuses the knowledge from the seed data.

\subsection{Ablations\label{sec:ablations}}

We conduct four ablation studies, as presented in Table~\ref{tab:ablation}.

\begin{table}[t]
    \centering
    \small
    \setlength{\tabcolsep}{1mm}
    \begin{tabular}{l|l|ccc}
        \toprule
        & \textbf{Dataset} & \textbf{MMLU} & \textbf{DROP} & \textbf{AVG.} \\
        \midrule
        \textbf{Base} & \ \ BoostQA & 64.78 & 47.41 & 51.66 \\
        \midrule
        \textbf{Refine} & - w/o Refine & 63.83 & 45.89 & 49.76 \\
        \midrule
        \textbf{Question} & - Multiple-Choice only & 64.20 & 46.32 & 51.26 \\
        \textbf{Type} & - Essay-Question only & 63.28 & 49.33 & 50.38 \\
        \midrule
        \textbf{Synthesis} & - Multi-Grade only & 65.15 & 46.42 & 50.38\\
        \textbf{Type} & - High-Difficulty only & 63.37 & 47.61 & 49.76 \\
        \midrule
        \textbf{General} & - w/ N-CC Document & 61.81 & 43.08 & 48.06 \\
        \textbf{Corpora} & - w/ FineWeb-Edu & 63.68 & 45.91 & 49.91 \\
        \bottomrule
    \end{tabular}
    \caption{Ablation results show the performance on MMLU, DROP, and the average across 12 benchmarks.}
    \label{tab:ablation}
\end{table}

\paragraph{Refinement enhances answer accuracy and improves model performance.} As shown in Table~\ref{tab:ablation}, implementing refinement results in an average improvement of 1.90\%. Our sampling indicates a 16.18\% rate of inconsistent answers before and after refinement, highlighting its crucial importance for data quality. We distill a model using DeepSeek-V3 for refinement, and model comparisons are available in Appendix~\ref{sec:appendix_findings_refine}. All subsequent experiments use the refined outputs.

\paragraph{Integrating both multiple-choice and essay question formats achieves optimal performance.} Table~\ref{tab:ablation} demonstrates the benefits of the multiple-choice format on benchmarks like MMLU, due to format alignment, while essay questions are vital for complex understanding tasks such as DROP. Combining both formats maximizes effectiveness, with multiple-choice questions facilitating efficient knowledge verification, and essay questions enabling deep comprehension, leading to synergistic improvements.

\paragraph{Integration of multi-grade and high-difficulty synthesis results in optimal performance.} As revealed in Table~\ref{tab:ablation}, multi-grade synthesis optimizes breadth and performs well on benchmarks requiring extensive knowledge, such as MMLU, whereas high-difficulty synthesis enhances depth for complex reasoning tasks, like DROP. Their integration overcomes individual limitations through complementary knowledge foundations and increased challenges.

\paragraph{\label{sec:ablation_blended_corpora}BoostQA maintains robustness through consistent improvements when mixed with various general corpora.} Table~\ref{tab:ablation} demonstrates that the base configuration of BoostQA delivers top performance. As shown in Table~\ref{tab:mainresult} and Table~\ref{tab:ablation}, when blended with alternative corpora, BoostQA with N-CC Document outperforms the original N-CC by an average of 3.15\%, and BoostQA with FineWeb-Edu outperforms the original FineWeb-Edu by 7.32\%. This corpora-agnostic enhancement effect affirms BoostQA's efficacy in overcoming structural and knowledge limitations across diverse corpora, serving as complementary data streams.

\subsection{Case Study}
We analyze the discipline and difficulty distribution of BoostQA in Appendix~\ref{sec:appendix_analysis}. Additionally, we compare the difficulty distribution between BoostQA and N-CC in Appendix~\ref{sec:appendix_findings_diffcmpwNCC}. To evaluate the quality of BoostQA, we conduct a case study, which is detailed in Appendix~\ref{sec:appendix_case}. Our findings show that the QA pairs generated using different seed formats, synthesizers, disciplines, and difficulty levels reveal a multidimensional diversity within our synthesis pipeline.

\section{Related Work}
Existing research establishes fundamental insights into the utilization of large-scale data for LLM training. General high-quality corpora, such as FineWeb-Edu~\citep{penedo2024fineweb}, offer broad linguistic coverage but often lack structured knowledge scaffolding. In contrast, QA data demonstrates superior capability, particularly in mid-training~\citep{wang2025octothinkermidtrainingincentivizesreinforcement, olmo20252olmo2furious}. \citet{maini2024rephrasing} generates supplementary QA pairs by rephrasing pre-training corpora. Moreover, \citet{cheng2024instruction} designs synthesis methodologies to enhance the quality of QA supervision signals. \citet{jiang2025mixcpt} investigates the performance of QA data incorporation in continual pre-training, while \citet{wang2025octothinkermidtrainingincentivizesreinforcement} examines its impact during mid-training.

Synthesis pipelines for QA data have evolved in diverse ways. \citet{su2024nemotron} leverages \ac{LLMs} to synthesize diverse QA formats from high-quality documents, yielding a 499.5B-token dataset of document-QA pairs, with QA data constituting merely one-tenth of the total. \citet{akter2025mind} employs role-specific prompting to generate a 138B-token mathematical dialogue dataset from corpora, which demonstrates lower content density than standard QA. \citet{zhou2025megamath} extracts and refines questions from high-quality web pages to produce a 7B-token synthetic mathematical QA dataset. Additionally, \citet{zhou2024jiuzhang} creates a knowledge distillation dataset to train a small LLM, synthesizing a 4.6B-token mathematical QA dataset for pre-training. Current methods struggle with data scalability, reusability, and cross-domain diversity. Our framework addresses these gaps by generating BoostQA, a large-scale, plug-and-play QA dataset, achieving SOTA results on 12 benchmarks.

\section{Conclusion}

In this paper, we identify deficiencies in model performance related to STEM disciplines and high-difficulty tasks through probe experiments. We propose a novel diversified synthesis pipeline that incorporates diverse seed curation, two-way synthesis of STEM-focused and high-difficulty data, and refinement. We have dedicated substantial resources to developing BoostQA, a plug-and-play, 100B-token large-scale synthetic QA dataset. Rigorous and comprehensive experiments validate BoostQA's 12.74\% average improvement on MMLU and CMMLU, with robust scalability across model size, data volume, and initial FLOPs.

\bibliography{aaai2026}

\appendix
\renewcommand{\thefigure}{A\arabic{figure}} 
\renewcommand{\thetable}{A\arabic{table}} 
\setcounter{figure}{0}
\setcounter{table}{0}

\section{Experimental Details \label{sec:appendix_train}}

\subsection{Training Details \label{sec:appendix_train_details}}
We fine-tune a synthesis model distilled from DeepSeek-R1 and an answer refinement model from DeepSeek-V3, with both models setting temperature to 0.6, top-p value to 0.95, and top-k to -1. All computations are executed on a dedicated cluster of 300 H20-141G GPUs.

We use 256 Ascend 910B NPUs to mid-train Llama-3 8B from the 2T-token checkpoint using 40B tokens of $\mathcal{D}_{\text{mt}}$, with each model training for over 22 hours. We implement it via the Megatron framework, optimized by the Adam algorithm with standard $\beta_1=0.9$ and $\beta_2=0.95$ parameters. The training employs a global batch size of 960 and a linearly decaying learning rate schedule initialized at $1.9\times 10^{-4}$ and terminating at $1.9\times 10^{-5}$.

In the model size scale experiment, we also test the performance of the dataset on Llama-3 1.7B and 16B with the same settings as the 8B model. For Llama-3 1.7B, we use 80 NPUs for training, with each model training for over 38 hours. For Llama-3 16B, we use 480 NPUs for training, with each model training for over 21 hours. Table~\ref{tab:appendix_train_hyper} presents the model configuration of Llama-3 1.7B, 8B, and 16B.

\begin{table}[ht]
    \centering
    \small
    \begin{tabular}{l|c|c|c}
        \toprule
        \textbf{Hyperparameter}	& \textbf{1.7B} & \textbf{8B} & \textbf{16B} \\
        \midrule
        Precision	& bfloat16 & bfloat16 & bfloat16 \\
        Layers	& 24 & 32 & 40 \\
        Hidden Size	& 2048 & 4096 & 5120 \\
        Attention Heads & 32 & 32 & 64 \\
        Head Type	& GQA & GQA & GQA \\
        Intermediate Size	& 8192 & 14336 & 18432 \\
        Vocab Size	& 131072 & 131072 & 163840 \\
        Sequence Length	& 8192 & 8192 & 8192 \\
        Activation	& SiLU & SiLU & SiLU \\
        Position Embedding	& RoPE & RoPE & RoPE \\
        \bottomrule
    \end{tabular}
    \caption{Model structure of Llama-3 1.7B, 8B, and 16B.}
    \label{tab:appendix_train_hyper}
\end{table}

\subsection{Datasets \label{sec:appendix_datasets}}
We compare BoostQA with large-scale QA datasets of different types and from different sources, as detailed in Table~\ref{tab:appendix_datasets}. For Nemotron-CC (N-CC), we maintain its original 9:1 document-to-QA ratio during experiments. We decompose it into N-CC Document and N-CC QA components. To isolate document effects, we substitute N-CC Document with KnowEdu as an alternative experimental setting and report optimal configurations as the result of N-CC.
\begin{table}[ht]
    \centering
    \setlength{\tabcolsep}{1mm}
    \small
    \begin{tabular}{l|ccccc}
        \toprule
        \textbf{Dataset} & \textbf{Synthesis}	& \textbf{CoT} &\textbf{Type}	&\textbf{Domain}	&\textbf{Tokens} \\
        \midrule
        N-CC		&\CheckmarkBold &\XSolidBrush & Doc. + QA	& General	& 499.5B \\
        N-CC QA &\CheckmarkBold &\XSolidBrush & QA & General & 51B \\
        YulanQA		&\XSolidBrush &\CheckmarkBold & QA	& General	& 4.92B \\ 
        N-MIND		&\CheckmarkBold &\CheckmarkBold & Conv.	& Math	& 138B \\
        MegaMathQA		&\CheckmarkBold &\CheckmarkBold & QA	& Math	& 7.0B \\
        JiuZhang3.0		&\CheckmarkBold &\CheckmarkBold & QA	& Math	& 4.6B \\
        \midrule
        BoostQA		&\CheckmarkBold &\XSolidBrush & QA	& General	& 100B \\
        BoostQA$_\text{CoT}$ &\CheckmarkBold &\CheckmarkBold & QA & General & 140B \\
        \bottomrule
    \end{tabular}
    \caption{Comparison with large-scale QA datasets. BoostQA$_\text{CoT}$ extends BoostQA by incorporating supplementary math CoT data BoostQA$_\text{MathCoT}$. In practice, the complete CoT dataset can contain significantly more tokens.}
    \label{tab:appendix_datasets}
\end{table}

\subsection{Scaling Details \label{sec:appendix_scaling}}
The specific accuracy values of the final checkpoint in the scaling experiments are presented in Table~\ref{tab:appendix_scaling}.

\begin{table}[ht]
    \centering
    \setlength{\tabcolsep}{1mm}
    \small
    \begin{tabular}{l|l|cccc}
        \toprule
        \textbf{Scale} & \textbf{Settings} & \textbf{MMLU}	&\textbf{CMMLU}	&\textbf{C-Eval}	&\textbf{STEM} \\
        \midrule
        & 1.7B: KnowEdu	& 45.32	& 48.95	& 45.39	& 38.63\\     
        & 1.7B: BoostQA	& 52.49	& 55.92	& 54.01	& 47.08 \\
        \textbf{Model} & 8B: KnowEdu & 63.53	& 68.08	& 67.18	& 53.86 \\
        \textbf{Size} & 8B: BoostQA	& 68.26	& 72.16	& 71.32	& 61.28 \\
        & 16B: KnowEdu & 68.61 & 72.53 & 69.23 & 61.44 \\
        & 16B: BoostQA & 72.34 & 76.32 & 73.77 & 66.21 \\
        \midrule
        & 40B: KnowEdu	& 58.17	& 62.99	& 61.98	& 49.16 \\
        \textbf{Data} & 40B: BoostQA	& 64.78	& 68.00	& 66.65	& 57.71 \\
        \textbf{Volume} & 190B: KnowEdu	& 60.72	& 67.12	& 65.07	& 51.11 \\
        & 190B: BoostQA	& 70.63	& 73.66	& 73.76	& 66.74 \\
        \midrule
        & 2T: KnowEdu	& 58.17	& 62.99	& 61.98	& 49.16 \\
        \textbf{Initial} & 2T: BoostQA	& 64.78	& 68.00	& 66.65	& 57.71 \\
        \textbf{FLOPs} & 10T: KnowEdu	& 63.53	& 68.08	& 67.18	& 53.86 \\
        & 10T: BoostQA	& 68.26	& 72.16	& 71.32	& 61.28 \\
        \bottomrule
    \end{tabular}
    \caption{Accuracy of the final checkpoint in the scaling experiments. Abbreviations: STEM = MMLU-STEM.}
    \label{tab:appendix_scaling}
\end{table}

\subsection{Analysis for BoostQA \label{sec:appendix_analysis}}

\begin{figure}[t]
\centering
\begin{subfigure}{0.48\linewidth}
    \centering
    \includegraphics[width=\linewidth, height=4cm, keepaspectratio]{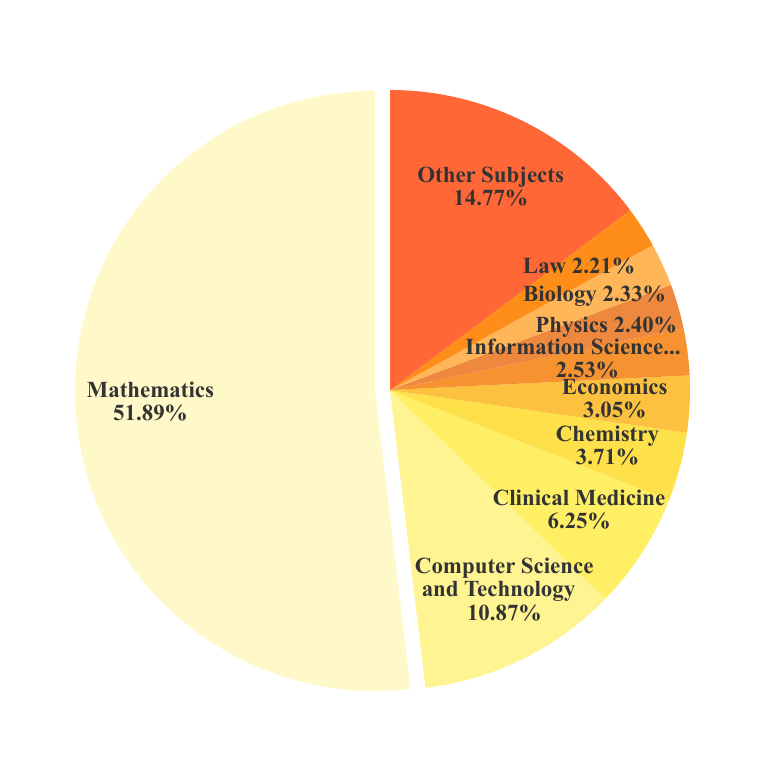}
    \caption{Discipline distribution.}
    \label{fig:subdis}
\end{subfigure}
\hfill
\begin{subfigure}{0.48\linewidth}
    \centering
    \includegraphics[width=\linewidth, height=4cm, keepaspectratio]{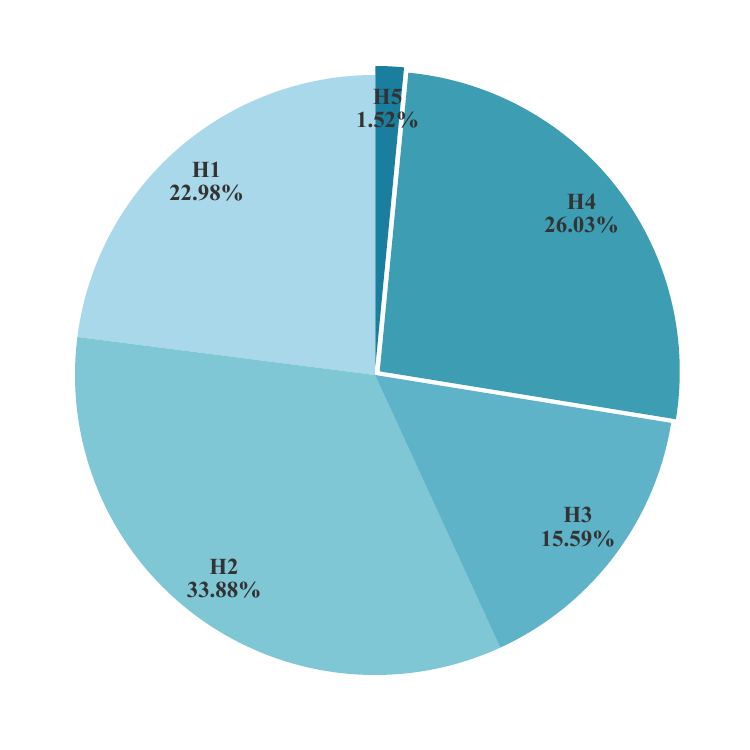}
    \caption{Difficulty distribution.}
    \label{fig:diffdis}
\end{subfigure}

\caption{The distribution of BoostQA.}
\label{fig:distribution}
\end{figure}

We sample instances from BoostQA to investigate its distinct discipline and difficulty stratification patterns, as shown in Figure~\ref{fig:distribution}. Discipline sampling reveals a pronounced concentration in STEM domains, with mathematics constituting 51.89\% of total content. Computer Science and Technology, and Clinical Medicine follow at 10.87\% and 6.25\% respectively, collectively establishing STEM dominance. This discipline skew demonstrates intentional synthesis prioritization rather than random distribution, strategically addressing capability gaps in complex analytical domains where traditional pre-training corpora exhibit insufficient coverage. Difficulty analysis demonstrates hierarchical stratification across the H1-H5 difficulty spectrum. Basic-level H1/H2 predominates at 56.86\% prevalence, establishing foundational knowledge scaffolding essential for progressive learning trajectories. Intermediate tiers H3 occupy 15.59\%, forming critical bridges between basic and extreme levels. While H4/H5 proportion increases to 27.55\%, as quantitative evidence of our synthesis module's capacity to amplify high-difficulty proportion.

\section{Findings \label{sec:appendix_findings}}

\subsection{Educational Stage Alignment \label{sec:appendix_findings_edu}}

We randomly sample 10,000 instances from each educational stage (high school, college, graduate) within the multi-grade synthetic dataset and employ DeepSeek-V3 to reassess their stage alignment. The distribution results, shown in Table~\ref{tab:findings_stage}, reveal an automatic decline in the educational stage alignment of synthetic data: high school-targeted requests predominantly yield junior high school content, university-targeted requests generate high school-level output, and graduate-level requests produce university-tier data. Consequently, achieving target-level question generation necessitates instructing the model to synthesize higher-tier content. For example, specifying graduate-level prompts to obtain university-level questions.

\begin{table}[htbp]
    \centering
    \setlength{\tabcolsep}{1mm}
    \small
    \begin{tabular}{c|c|cccccc}
        \toprule
        \textbf{Targeted} & \textbf{Match} & \multicolumn{6}{c}{\textbf{Actual Grade Distribution}}\\
        \textbf{Stage} & \textbf{Acc.} & prim. & junior. & high. & coll. & grad. & other \\
        \midrule
        high. & 30.31 & 2.31 & \textbf{55.60} & 0.03 & \underline{38.62} & 0.10 & 3.34 \\
        coll. & 42.87 & 1.40 & \underline{36.71} & \textbf{56.41} & 0.16 & 0.59 & 4.73 \\
        grad. & 7.86 & 0.63 & 18.70 & \underline{24.93} & \textbf{52.38} & 0.01 & 3.35 \\
        \bottomrule
    \end{tabular}
    \caption{Stage alignment testing for multi-grade synthesis. The largest and second largest proportions are in bold and underlined, respectively. Abbreviations: prim. = primary school, junior. = junior high school, high. = high school, coll. = college, grad. = graduate.}
    \label{tab:findings_stage}
\end{table}

\subsection{Difficulty Distribution Comparison of Synthetic Datasets\label{sec:appendix_findings_diffcmpwNCC}}

\begin{figure}[ht]
\centering
\includegraphics[width=0.9\columnwidth]{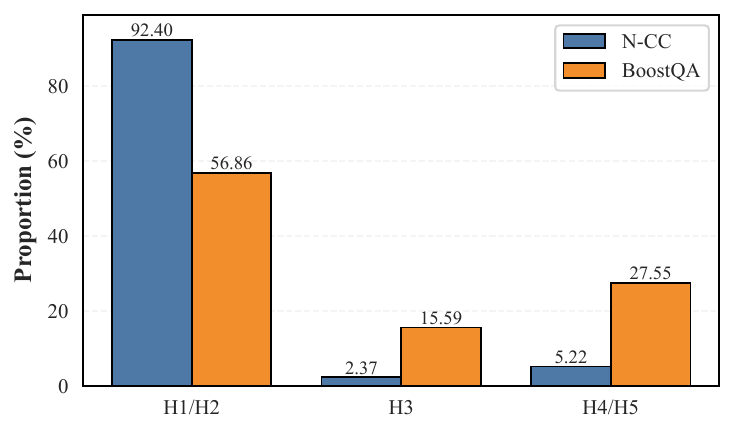} 
\caption{Difficulty distribution comparison between N-CC and BoostQA.}
\label{fig:diffcmpwNCC}
\end{figure}
We compare the difficulty distribution of BoostQA with N-CC, another general synthetic QA dataset. As illustrated in Figure~\ref{fig:diffcmpwNCC}, BoostQA demonstrates a significant increase in the high-difficulty H4/H5 proportion.

\subsection{Comparison of Models Used for Refinement \label{sec:appendix_findings_refine}}

We test the performance of the synthetic datasets refined by different models. The results in Table~\ref{tab:findings_refine} show that the average effects of the three models are similar.

\begin{table}[ht]
    \centering
    \small
    \setlength{\tabcolsep}{1mm}
    \begin{tabular}{l|c|c|c}
        \toprule
        \textbf{Metric} & \textbf{V3} & \textbf{R1 w/o think} & \textbf{R1 w/ think} \\
        \midrule
        \textbf{MMLU}      & 64.78 & 64.77 & 65.01 \\
        \textbf{CMMLU}     & 68.00 & 67.78 & 67.43 \\
        \textbf{GSM8K}     & 47.73 & 48.79 & 49.35 \\
        \textbf{MATH}      & 14.00 & 13.50 & 12.00 \\
        \textbf{HellaSwag} & 60.50 & 62.00 & 66.00 \\
        \textbf{WinoGrande} & 44.00 & 45.00 & 44.00 \\
        \textbf{ARC-C}     & 80.00 & 78.00 & 77.00 \\
        \textbf{AVG.}      & 54.14 & 54.26 & 54.40 \\
        \bottomrule
    \end{tabular}
    \caption{Comparison of datasets refined by different DeepSeek models.}
    \label{tab:findings_refine}
\end{table}

\subsection{Performance on MMLU Subsets \label{appendix_findings_mmlusubsets}}

\begin{table}[ht]
    \centering
    \small
    \begin{tabular}{lccc}
        \toprule
        \textbf{Dataset} & \textbf{Social} & \textbf{Humanity} & \textbf{Other} \\
        \midrule
        Pre-training & 63.85 & 59.44 & 58.26 \\
        FineWeb-Edu & 63.50 & 60.48 & 58.67 \\
        KnowEdu & 67.20 & \underline{62.44} & 60.32 \\
        \midrule
        N-CC & 67.31 & 61.84 & \underline{61.88} \\
        YulanQA & 65.65 & 59.28 & 58.84 \\
        N-MIND & \underline{67.40} & 61.23 & 59.60 \\
        MegaMathQA & 63.02 & 58.39 & 59.18 \\
        JiuZhang3.0 & 65.19 & 59.80 & 58.37 \\
        \midrule
        BoostQA & \textbf{72.83} & \textbf{66.26} & \textbf{66.40} \\
        \bottomrule
    \end{tabular}
    \caption{Comparison on MMLU subsets. The best and second best are in bold and underlined, respectively. }
    \label{tab:mmlusub}
\end{table}

The MMLU subset results shown in Table~\ref{tab:mmlusub} further substantiate BoostQA's cross-domain superiority, outperforming general corpora and other synthetic QA baselines across all categories: MMLU-social, MMLU-humanity, and MMLU-other. This consistent advantage persists despite structural divergence in knowledge representation, confirming that our synthesis pipeline fundamentally enhances conceptual generalization rather than optimizing domain-specific heuristics. Crucially, BoostQA generates transferable reasoning signals through its stem-focused educational diversified progression and difficulty booster. When integrated with the high-quality educational corpora, it establishes robust cross-discipline cognitive scaffolding unavailable in baselines. The observed performance patterns validate BoostQA's unique capacity to transcend discipline boundaries while maintaining task-agnostic robustness.

\onecolumn

\section{Annotation System \label{sec:appendix_annotation}}

The discipline classifier categorizes content into 62 first-level disciplines\footnote{GB/T 13745-2008 taxonomy}. To ensure labeling efficiency while maintaining quality, we implement a two-stage annotation pipeline. First, we use DeepSeek-R1 with the discipline-constrained prompt and generate preliminary labels for 20M seed samples. Then, through uniform stratified sampling across all 62 disciplines, we curate a balanced subset of 500K high-confidence samples to finetune Qwen2.5-7B-Instruct, yielding the specialized subject classifier. Empirical validation shows 82.18\% label consistency between our specialized classifier and DeepSeek-R1, confirming reliable knowledge distillation. The prompt used to annotate data with discipline and train the corresponding labeler is shown as follows.
\begin{tcolorbox}[colback=white!95!gray,colframe=gray!50!black,rounded corners,label={prompt-discipline-classifier}, title={Prompt for Discipline Classifier}]
\begin{lstlisting}[breaklines=true, xleftmargin=0pt, breakindent=0pt, columns=fullflexible, mathescape, numbers=none]
Act as an educational taxonomist. Classify the input question into our standardized discipline hierarchy using sequential reasoning, then output strictly in JSON format:
1. Primary Discipline Identification
   Select exactly one primary discipline from:  
   {Discipline List}  
   - Use "cross-discipline" only for explicit multi-domain integration  
   - Assign "Other" only if no discipline matches >=60% relevance  
2. Secondary Discipline Assignment 
   - Identify the most specific applicable sub-discipline 
   - Null if primary discipline has no sub-domains  
   - Use "General" for non-specialized content  
3. Validation Rules 
   - Reject non-educational content -> Output "Invalid"  
   - Correct spelling/terminology variations before classification  
Output Schema: 
{
  "primary_discipline": "",
  "secondary_discipline": "",
  "confidence": 0.0-1.0,
  "rejection_reason": null
}
Input: {Seed Data}
\end{lstlisting}
\end{tcolorbox}

The list of 62 primary disciplines is as follows.

\begin{tcolorbox}[colback=white!95!gray,colframe=gray!50!black,rounded corners,label={prompt-discipline-list}, title={Discipline List (62)}]
\begin{lstlisting}[breaklines=true, xleftmargin=0pt, breakindent=0pt, columns=fullflexible, mathescape, numbers=none]
['Mathematics', 'Computer Science and Technology', 'Clinical Medicine', 'Chemistry', 'Economics', 'Information Science and Systems Science', 'Physics', 'Biology', 'Law', 'Philosophy', 'Sociology', 'Literature', 'Psychology', 'Statistics', 'History', 'Power and Electrical Engineering', 'Earth Science', 'Management Science', 'Electronics and Communication Technology', 'Linguistics', 'Preventive Medicine and Public Health', 'Political Science', 'Education Science', 'Aerospace Science and Technology', 'Astronomy', 'Materials Science', 'Mechanics', 'Sports Science', 'Ethnology and Cultural Studies', 'Basic Medicine', 'Environmental Science and Resource Science', 'Journalism and Communication', 'Religious Studies', 'Engineering and Technology Related to Information and Systems Science', 'Food Science and Technology', 'Engineering and Technology', 'Art Studies', 'Mechanical Engineering', 'Traditional Chinese Medicine and Chinese Materia Medica', 'Pharmacy', 'Civil and Architectural Engineering', 'Chemical Engineering', 'Nuclear Science and Technology', 'Marxism', 'Agronomy', 'Energy Science and Technology', 'Transportation Engineering', 'Military Science', 'Safety Science and Technology', 'Animal Husbandry and Veterinary Science', 'Archaeology', 'Engineering and Technology Related to Product Applications', 'Library, Information and Documentation Science', 'Geomatics Science and Technology', 'Aquaculture Science', 'Metallurgical Engineering Technology', 'Hydraulic Engineering', 'Military Medicine and Special Medicine', 'Textile Science and Technology', 'Mining Engineering Technology', 'Forestry', 'Engineering and Technology Related to Natural Sciences']
\end{lstlisting}
\end{tcolorbox}

Parallelly, the difficulty scorer operationalizes human performance metrics, defined as pass rates under standardized one-hour testing conditions with QS Top 100 university students majoring in relevant disciplines, to calibrate five difficulty tiers (H1-H5). We implement a multi-stage annotation pipeline. Initial difficulty annotations are generated by DeepSeek-R1 through a structured prompt that simulates human problem-solving behaviors. This prompt-based simulation phase produces preliminary difficulty estimates for 500K QA pairs. Subsequently, we distill this knowledge by training Qwen2.5-14B-Instruct from the annotated data. Validation shown through expert assessment (five PhD evaluators, Krippendorff’s $\alpha$ = 0.85) confirms a strong correlation (Pearson’s $r$ = 0.92) between model predictions and actual human performance metrics. The prompt used to annotate data with difficulty labels and train the corresponding labeler is shown as follows.

\begin{tcolorbox}[colback=white!95!gray,colframe=gray!50!black,rounded corners,label={prompt-difficulty-scorer}, title={Prompt for Difficulty Scorer}]
\begin{lstlisting}[breaklines=true, xleftmargin=0pt, breakindent=0pt, columns=fullflexible, mathescape, numbers=none]
Act as an educational assessment expert, analyze the provided question through sequential reasoning and output strictly in JSON format: 
1. Knowledge Analysis
   - Core concepts (<=3): [comma-separated list]  
   - Integration type: {single-concept | cross-chapter | cross-discipline}  
2. Cognitive Tier (Bloom's Taxonomy)  
   {memory | understanding | application | analysis | synthesis | evaluation}  
3. Difficulty Assessment
   - Estimated pass rate (P) for QS Top 100 university majors: [0-100%]  
   - Tier:  
     - extreme: P < 10%  
     - challenge: 10% <= P < 30%  
     - improvement: 30% <= P < 50%  
     - standard: 50% <= P < 80%  
     - basic: P >= 80%  
     - other: invalid inputs  
4. Exception Handling
   - Mark "other" for non-questions/unanswerable items  
   - Correct minor errors (e.g., missing correct options) before assessment  
   - Ignore provided solutions/answers 
Output Schema:  
{
  "difficulty_tier": "basic|standard|improvement|challenge|extreme|other",
  "rationale": [
    "Involves {N} core knowledge points",
    "Cognitive level: {Bloom's tier}",
    "Estimated pass rate: approximately {XX}% for target cohort"
  ]
}
Input: {Seed Data}
\end{lstlisting}
\end{tcolorbox}

\section{Synthesis Prompts \label{sec:appendix_prompts_synthesis}}
The prompts and specific rules used to synthesize diverse knowledge-intensive QA pairs are as follows.
\begin{tcolorbox}[colback=white!95!gray,colframe=gray!50!black,rounded corners,label={prompt-multi-grade}, title={Prompt for Multi-Grade Synthesizer}]
\begin{lstlisting}[breaklines=true, xleftmargin=0pt, breakindent=0pt, columns=fullflexible, mathescape, numbers=none]
Act as a {Role Assigner} educator, analyze the knowledge points assessed by the provided {Seed Format}. Generate {Number} novel questions adhering to these requirements:

1. Questions must demonstrate substantial differentiation while testing application of identified knowledge points. 
2. Difficulty must align with {Role Assigner} standards through:  
   a) Down-scaling overqualified knowledge points to prerequisite concepts;
   b) Up-scaling underqualified points to advanced applications.
3. Linguistic consistency must be maintained with the input question.
{Format-specific Constraints}

Output Schema: {Format-specified JSON}
Input: {Seed Data}
\end{lstlisting}
\end{tcolorbox}
\{Role Assigner\} can be assigned as high school, college, and graduate. In our experiment, \{Number\} is usually set to 10. \{Format-specific Constraints\} and \{Format-specific JSON\} are controlled by the rule enforcer and vary depending on whether the targeted synthetic question type is multiple-choice or essay question format, following the specific rules below:

\begin{tcolorbox}[colback=white!95!gray,colframe=gray!50!black,rounded corners,label={prompt-rules}, title={Rules for Targeted Synthetic Qustion Type}]
\begin{lstlisting}[breaklines=true, xleftmargin=0pt, breakindent=0pt, columns=fullflexible, mathescape, numbers=none]
Format-specific Constraints:
Multiple-Choice: 4. The generated question type is multiple-choice. For each question, four alternative options must be generated, and among the four options, there must be one correct answer.
Essay-question: 4. The generated question type is essay-question. For each question, the solution steps and the final correct answer are provided. The generated questions cannot be open-ended questions (such as those of the solution type, thinking type, information listing type, etc.), but must be self-contained with a final answer that can be determined as correct.

Format-specified JSON:
Multiple-Choice: [{"question": "", "options": [],  "answer_index": 0-3}, ...]
Essay Question: [{"question": "", "solution": "",  "answer": ""}, ...]

\end{lstlisting}
\end{tcolorbox}

The following prompt for high-difficulty synthesizer is similar to the multi-grade one but specifies the role as graduate and adds a difficulty booster part.

\begin{tcolorbox}[colback=white!95!gray,colframe=gray!50!black,rounded corners,label={prompt-high-difficulty}, title={Prompt for High-Difficulty Synthesizer}]
\begin{lstlisting}[breaklines=true, xleftmargin=0pt, breakindent=0pt, columns=fullflexible, mathescape, numbers=none]
Act as a graduate educator, analyze the knowledge points assessed by the provided {Seed Format}. Generate {Number} novel questions adhering to these requirements:

1. Questions must demonstrate substantial differentiation while testing application of identified knowledge points  
2. Difficulty must align with {Role Assigner} standards through:  
   a) Down-scaling overqualified knowledge points to prerequisite concepts  
   b) Up-scaling underqualified points to advanced applications  
3. Linguistic consistency must be maintained with the input question
{Format-specific Constraints}

[Difficulty Booster]
1. Knowledge Analysis:  
   - Core concepts (<=3)  
   - Integration type: {single | cross-chapter | cross-discipline}  
2. Cognitive Tier (Bloom's Taxonomy):  
   {memory | understanding | application | analysis | synthesis | evaluation}  
3. Difficulty Validation:  
   - Estimate pass rate 0 <= P <= 100%
   - Tier:  
     - extreme: P < 10%  
     - challenge: 10% <= P < 30%  
     - improvement: 30% <= P < 50%  
     - standard: 50% <= P < 80%  
     - basic: P >= 80%   
   - REJECT if not challenge/extreme (P >= 30%) 

Output Schema: {Format-specified JSON}
Input: {Seed data}
\end{lstlisting}
\end{tcolorbox}

\section{Case Study \label{sec:appendix_case}}

\begin{longtable}{p{0.48\textwidth}|p{0.48\textwidth}}
\caption{Case of different seed and different synthesizer.}
\label{tab:case_seed_syn} \\
\toprule
\midrule
\endfirsthead

\caption[]{(continued)} \\
\toprule
\midrule
\endhead

\midrule
\endfoot

\bottomrule
\endlastfoot

\small \textbf{Seed Data 1 (High-quality Web Page)} & \small \textbf{Seed Data 2 (QA)} \\
\midrule
\small To divide a 4-digit number by a single-digit number, follow these steps: 1. Divide the first digit of the 4-digit number by the single-digit number. If the result is less than the single-digit number, consider the first two digits. 2. Multiply the result by the single-digit number and subtract it from the considered digits. 3. Bring down the next digit and repeat the process until all digits are considered. 4. The final result will be the quotient. Examples: ...  & \small How does Nuclear Magnetic Resonance Imaging (MRI) work, and what factors affect the signal produced by the nuclei in the magnetic field? Answer: Nuclear Magnetic Resonance Imaging (MRI) is a non-invasive medical imaging technique that uses magnetic fields and radio waves to generate detailed images of internal body structures. It relies on the principles of nuclear magnetic resonance (NMR), primarily targeting hydrogen nuclei (protons) in water and fat molecules due to their abundance in biological tissues. Below is a step-by-step explanation of how MRI works, followed by key factors affecting the signal. \\
\midrule
\small \textbf{Synthetic QA pairs (multi-grade, multiple-choice, n = 10)} & \small \textbf{Synthetic QA pairs (multi-grade, multiple-choice, n = 10)} \\
\midrule
\small \textbf{Question 1}: When dividing a 4-digit number by a single-digit number, what should you do if the first digit is smaller than the divisor? A. Consider only the first digit\ \ \ \ B. Consider the first two digits\ \ \ \ C. Skip to the next digit immediately\ \ \ \ D. Add the divisor to the first digit 
& \small \textbf{Question 1}: In MRI, what is the primary function of the gradient magnetic field during slice selection?  A. To generate T1-weighted contrast\ \ \ \ B. To spatially encode the position of nuclei along the z-axis\ \ \ \ C. To amplify the RF excitation pulse\ \ \ \ D. To reduce T2* decay artifacts \\
\small \textbf{Answer 1}: B. Consider the first two digits & \small \textbf{Answer 1}: B. To spatially encode the position of nuclei along the z-axis \\
\small \textbf{Question 2}: What must always be true about the remainder in division? A. Remainder equals the divisor\ \ \ \ B. Remainder is greater than the divisor\ \ \ \ C. Remainder is less than the divisor\ \ \ \ D. Remainder must be zero 
& \small \textbf{Question 2}: Which tissue property directly determines proton density-weighted image contrast? A. T1 relaxation time\ \ \ \ B. T2 relaxation time\ \ \ \ C. Concentration of hydrogen nuclei\ \ \ \ D. Magnetic susceptibility \\
\small \textbf{Answer 2}: C. Remainder is less than the divisor 
& \small \textbf{Answer 3}: C. Concentration of hydrogen nuclei\\
\small \textbf{Question 3}: Which equation correctly verifies division results? A. Divisor × Quotient = Dividend\ \ \ \ B. Divisor × Remainder = Dividend\ \ \ \ C. Divisor × Quotient + Remainder = Dividend\ \ \ \ D. Quotient × Remainder = Dividend 
& \small \textbf{Question 3}: Why does cerebrospinal fluid (CSF) appear bright in T2-weighted MRI images? A. Short T1 relaxation time\ \ \ \ B. High proton density\ \ \ \ C. Long T2 relaxation time\ \ \ \ D. Low diffusion coefficient \\
\small \textbf{Answer 3}: C. Divisor × Quotient + Remainder = Dividend 
& \small \textbf{Answer 3}: C. Long T2 relaxation time \\
\small \textbf{Question 4}: In the division process, after multiplying and subtracting, what do you bring down next? A. The previous remainder\ \ \ \ B. The divisor\ \ \ \ C. The next digit of the dividend\ \ \ \ D. The quotient digit 
& \small \textbf{Question 4}: What is the primary source of T2* decay in gradient-echo sequences? A. Spin-spin interactions\ \ \ \ B. Magnetic field inhomogeneities\ \ \ \ C. RF pulse imperfections\ \ \ \ D. Gradient coil heating \\
\small \textbf{Answer 4}: C. The next digit of the dividend 
& \small \textbf{Answer 4}: B. Magnetic field inhomogeneities \\
\small \textbf{Question 5}: What is the first step when calculating 5,274 ÷ 6? A. Divide 5 by 6\ \ \ \ B. Divide 52 by 6\ \ \ \ C. Divide 527 by 6\ \ \ \ D. Divide 5274 by 6  
& \small \textbf{Question 5}: If an MRI image exhibits 'zipper artifacts' along the frequency-encoding direction, what is the most likely cause? A. Insufficient TR time\ \ \ \ B. RF interference from external sources\ \ \ \ C. Gradient amplifier failure\ \ \ \ D. Patient motion during scan \\
\small \textbf{Answer 5}: B. Divide 52 by 6 
& \small \textbf{Answer 5}: B. RF interference from external sources  \\
\small ... & \small ... \\

\midrule

\small \textbf{Synthetic QA pairs (high-difficulty, multiple-choice, n = 10)} 
& \small \textbf{Synthetic QA pairs (high-difficulty, multiple-choice, n = 10)} \\
\midrule
\small \textbf{Question 1}: In a modified division algorithm where remainders are allowed to be negative, when dividing 4107 by 6, what would be the smallest absolute value remainder possible? A. 3\ \ \ \ B. -3\ \ \ \ C. 2\ \ \ \ D. -2 
& \small \textbf{Question 1}: In MRI imaging, when comparing T1-weighted and T2-weighted images, which parameter combination primarily determines the contrast difference? A. TR=500ms, TE=20ms\ \ \ \ B. TR=3000ms, TE=100ms\ \ \ \ C. TR=2000ms, TE=20ms\ \ \ \ D. TR=500ms, TE=80ms \\
\small \textbf{Answer 1}: B. -3 
& \small \textbf{Answer 1}: B. TR=3000ms, TE=100ms \\
\small \textbf{Question 2}: Given the division 7608 ÷ 8 = 951, if a single-digit error occurs in the quotient's tens place during calculation, what maximum absolute difference could this error create in the final result? A. 80\ \ \ \ B. 40\ \ \ \ C. 800\ \ \ \ D. 8 
& \small \textbf{Question 2}: For a spin-echo sequence, what happens to the MR signal if both TR and TE are doubled while maintaining other parameters? A. T1 weighting increases\ \ \ \ B. Signal-to-noise ratio improves\ \ \ \ C. T2 weighting dominates\ \ \ \ D. Specific absorption rate decreases \\
\small \textbf{Answer 2}: A. 80 
& \small \textbf{Answer 2}: C. T2 weighting dominates \\
\small \textbf{Question 3}: Using the division principle from the examples, what is the minimum number of multiplication/verification steps required to divide any 4-digit number by a single-digit divisor? A. 4 steps\ \ \ \ B. 5 steps\ \ \ \ C. Depends on carries\ \ \ \ D. Always 3 steps 
& \small \textbf{Question 3}: Which factor has the MOST significant impact on the Larmor frequency in clinical MRI systems? A. Patient body temperature\ \ \ \ B. Main magnetic field strength\ \ \ \ C. Radiofrequency pulse amplitude\ \ \ \ D. Gradient coil performance \\
\small \textbf{Answer 3}: A. 4 steps 
& \small \textbf{Answer 3}: B. Main magnetic field strength \\
\small \textbf{Question 4}: For the division 6585 ÷ 5 = 1317, if the dividend is increased by k while keeping the same quotient, what is the maximum k such that the remainder doesn't change? A. 4\ \ \ \ B. 5\ \ \ \ C. 10\ \ \ \ D. 15 
& \small \textbf{Question 4}: When implementing fat suppression techniques, which physical property difference is primarily exploited? A. T1 relaxation time\ \ \ \ B. Proton density\ \ \ \ C. Chemical shift\ \ \ \ D. Magnetic susceptibility \\
\small \textbf{Answer 4}: A. 4 
& \small \textbf{Answer 4}: C. Chemical shift \\
\small \textbf{Question 5}: For a cryptographic application using modulo 9 arithmetic, what is 3408 mod 3 equivalent to, given the original division example? A. 0\ \ \ \ B. 1\ \ \ \ C. 2\ \ \ \ D. 3 
& \small \textbf{Question 5}: What is the primary reason for using phase encoding gradients in MRI? A. Slice selection\ \ \ \ B. Frequency determination\ \ \ \ C. Spatial localization in one dimension\ \ \ \ D. Magnetic field homogenization \\
\small \textbf{Answer 5}: A. 0 
& \small \textbf{Answer 5}: C. Spatial localization in one dimension \\
\small ... & \small ...\\
\end{longtable}

\begin{longtable}{p{0.98\textwidth}}
\caption{Samples for synthetic QA pairs of different disciplines. }
\label{tab:case_discipline} \\
\toprule
\midrule
\endfirsthead

\caption[]{(continued)} \\
\toprule
\midrule
\endhead

\midrule
\endfoot

\bottomrule
\endlastfoot

\small \textbf{Mathematics} \\
\midrule
\small \textit{\textbf{Sample 1}:} Question: A machine fills 120 bottles in 2.5 hours. What is the average time to fill one bottle? A. 1.25 minutes\ \ \ \ B. 1.5 minutes\ \ \ \ C. 1.75 minutes\ \ \ \ D. 2.0 minutes \\
\small Answer: A. 1.25 minutes \\
\small \textit{\textbf{Sample 2}:} Question: In a modified deck with 13 cards per suit (52 total), what is the expected number of draws until three consecutive cards of the same suit appear? Round to the nearest tenth. \\
\small Answer: 401.4 \\
\midrule
\small \textbf{Computer Science and Technology} \\
\midrule
\small \textit{\textbf{Sample 1}:} Question: Which BEST describes the method resolution order (MRO) impact when using mixins in Python? A. Mixins always take precedence over base classes\ \ \ \ B. MRO follows C3 linearization rules\ \ \ \ C. Depth-first left-right search order\ \ \ \ D. Mixins are ignored in super() calls \\
\small Answer: B. MRO follows C3 linearization rules \\
\small \textit{\textbf{Sample 2}:} Question: Calculate a doubly linked list function that supports fuzzy deletion, which requires randomly deleting nodes based on a given probability distribution while maintaining the integrity of the remaining linked list structure. Please describe the selection and implementation methods of the probability model. \\
\small Answer: Implementation steps: 1) Select the probability model and assign node probabilities. 2) Randomly select nodes based on cumulative distribution. 3) Perform the standard doubly linked list deletion operation \\
\midrule
\small \textbf{Clinical Medicine} \\
\midrule
\small \textit{\textbf{Sample 1}:} Question: Which pathogen is classified as a virus in the given scenario? A. Giardia\ \ \ \ B. Trichinella\ \ \ \ C. Hepatitis A\ \ \ \ D. Salmonella enterica \\
\small Answer: C. Hepatitis A \\
\small \textit{\textbf{Sample 2}:} Question: Which three serum biomarkers would best differentiate pre-eclampsia-related pulmonary edema from septic ARDS in a postpartum patient? \\
\small Answer: sFlt-1, PlGF, and Procalcitonin. \\
\midrule
\small \textbf{Chemistry} \\
\midrule
\small \textit{\textbf{Sample 1}:} Question: A linear polymer is synthesized via condensation reactions using 20 monomers. How many water molecules are released? A. 19 water molecules\ \ \ \ B. 20 water molecules\ \ \ \ C. 18 water molecules\ \ \ \ D. 21 water molecules \\
\small Answer: A. 19 water molecules \\
\small \textit{\textbf{Sample 2}:} Question: Design a theoretical calculation scheme to determine whether $\text{O}_2$ adsorption on the catalyst surface is physical or chemical adsorption through quantum chemical methods, and list three key calculation parameters. \\
\small Answer: The key calculation parameters are: adsorption energy (E\_ads), charge density difference (CDD), and $\text{O}_2$ bond length variation. \\
\midrule
\small \textbf{Economics} \\
\midrule
\small \textit{\textbf{Sample 1}:} Question: If a depositor holds \$300,000 in a single checking account at an FDIC-insured bank, how much of this amount is covered by FDIC insurance? A. \$100,000\ \ \ \ B. \$250,000\ \ \ \ C. \$300,000\ \ \ \ D. \$500,000 \\
\small Answer: B. \$250,000 \\
\small \textit{\textbf{Sample 2}:} Question: Based on the mechanism of real estate mortgage loans, analyze the fundamental legal obstacles that make it difficult for developing countries to popularize mortgage loans and their second-order impact on the structure of the financial market. \\
\small Answer: The fundamental legal obstacles to the popularization of mortgage loans in developing countries mainly include the imperfect property rights registration system, the low efficiency of law enforcement and the difficulty in disposing of collateral. The second-order impact of these obstacles on the structure of the financial market is manifested as insufficient development of long-term financing tools, increased reliance on informal financial channels to increase systemic risks, and restrictions on the depth and breadth of the financial market. \\
\midrule
\small \textbf{Information Science and Systems Science} \\
\midrule
\small \textit{\textbf{Sample 1}:} Question: What is the primary technological factor that diminished RIM's corporate market advantage? A. Superior hardware design by competitors\ \ \ \ B. Improved security features in iOS devices\ \ \ \ C. Development of 5G network technology\ \ \ \ D. Reduction in smartphone production costs \\
\small Answer: B. Improved security features in iOS devices \\
\small \textit{\textbf{Sample 2}:} Question: In the detection of abnormal behaviors in social networks, how to construct an abnormal scoring function for user behavior sequences based on the hidden Markov model? \\
\small Answer: The anomaly scoring function for the user behavior sequence constructed based on the hidden Markov model is $S(X) = -\log P(X|\lambda)$, where $P(X|\lambda)$ is calculated through the forward algorithm. Parameter estimation is iteratively optimized using the Baum-Welch algorithm. \\
\midrule
\small \textbf{Physics} \\
\midrule
\small \textit{\textbf{Sample 1}:} Question: A train 200 m long crosses a man walking at 5 km/h in the same direction in 15 seconds. What is the train's speed? A. 45 km/h\ \ \ \ B. 50 km/h\ \ \ \ C. 53 km/h\ \ \ \ D. 58 km/h \\
\small Answer: C. 53 km/h \\
\small \textit{\textbf{Sample 2}:} Question: The object is moving at an initial velocity of 20m/s and decelerates uniformly. The displacement in the last second before it stops is 2m. Find the magnitude of the acceleration. \\
\small Answer: 4 \\
\midrule
\small \textbf{Biology} \\
\midrule
\small \textit{\textbf{Sample 1}:} Question: Which symbiotic relationship is characterized by one organism benefiting while the host is harmed but rarely killed? A. Mutualism\ \ \ \ B. Commensalism\ \ \ \ C. Parasitism\ \ \ \ D. Amensalism \\
\small Answer: C. Parasitism \\
\small \textit{\textbf{Sample 2}:} Question: Two pairs of alleles (X/x, Y/y) are inherited independently, with X being dominant and Y being dominant (the presence of Y masks the X phenotype). The homozygous parents are XXYY (Y dominant) and xxyy (non-dominant), and F$_1$ self-pollinates to F$_2$. Calculate the proportion of individuals with phenotypes different from those of their parents in F$_2$. \\
\small Answer: 3/16 \\
\midrule
\small \textbf{Law} \\
\midrule
\small \textit{\textbf{Sample 1}:} Question: A clinic uses an unencrypted email service to send lab results to patients. Which safeguard is most critically missing? A. Multi-factor authentication\ \ \ \ B. Business Associate Agreement\ \ \ \ C. Technical protection for ePHI in transit\ \ \ \ D. Annual HIPAA training \\
\small Answer: C. Technical protection for ePHI in transit \\
\small \textit{\textbf{Sample 2}:} Question: County Party Secretary A promised to change the land use for developer B, but demanded that B donate the bribe of 2 million yuan to the charity foundation designated by A. How to determine the nature of Party A's actions? \\
\small Answer: Party A's actions constitute the crime of accepting bribes, with the amount of the bribe being 2 million yuan. \\
\end{longtable}

\begin{longtable}{p{0.98\textwidth}}
\caption{Samples for synthetic QA pairs of different difficulty levels.}
\label{tab:case_difficulty} \\
\toprule
\midrule
\endfirsthead

\caption[]{(continued)} \\
\toprule
\midrule
\endhead

\midrule
\endfoot

\bottomrule
\endlastfoot

\small \textbf{H1} \\
\midrule
\small \textit{\textbf{Sample 1}:} Question: Which file extension is typically associated with pip executables on Windows? A. .py\ \ \ \ B. .sh\ \ \ \ C. .exe\ \ \ \ D. .dll \\
\small Answer: C. .exe \\
\small \textit{\textbf{Sample 2}:} Question: A regular octahedron has 6 vertices and 12 edges. A segment that joins two vertices not joined by an edge is called a diagonal. How many diagonals does a regular octahedron have? \\
\small Answer: 3 \\
\midrule
\small \textbf{H2} \\
\midrule
\small \textit{\textbf{Sample 1}:} Question: What is the primary environmental concern regarding offshore drilling in Arctic conditions? A. Air emissions from rig operations\ \ \ \ B. Ice-scouring damage to subsea infrastructure\ \ \ \ C. Drill cuttings dispersion in cold water\ \ \ \ D. Noise pollution affecting marine mammals \\
\small Answer index: B. Ice-scouring damage to subsea infrastructure \\
\small \textit{\textbf{Sample 2}:} Question: The chord length of the circle x² + y² =9 intersecting the line 4x +3y +c =0 is 6. Find the value of c. \\
\small Answer: 0 \\
\midrule
\small \textbf{H3} \\
\midrule
\small \textit{\textbf{Sample 1}:} Question: Which phenomenon best illustrates the transformation process described in the definition? A. A novelist copyrighting their manuscript\ \ \ \ B. A TikTok story becoming a published anthology\ \ \ \ C. Academic peer-review process\ \ \ \ D. Direct translation of sacred texts \\
\small Answer: B. A TikTok story becoming a published anthology \\
\small \textit{\textbf{Sample 2}:} Question: The pulley block connects objects with masses m1=3kg and m2=2kg. Find (a) the increase in system kinetic energy within 2 seconds after release; (b) the change in mechanical energy of m1. (Pulley mass and friction are disregarded, g=10m/s²) \\
\small Answer: (a) 40J; (b) -96J \\
\midrule
\small \textbf{H4} \\
\midrule
\small \textit{\textbf{Sample 1}:} Question: Which of the following expressions satisfies the property of being divisible by $3^{n+1}$ but not by $3^{n+2}$? A. $7^{2^n} -1$\ \ \ \ B. $4^{3^n} +1$\ \ \ \ C. $5^{2^n} -2$\ \ \ \ D. $2^{3^n} +1$ \\
\small Answer: D.$ 2^{3^n} +1$ \\
\small \textit{\textbf{Sample 2}:} Question: When the difference in the two electron integrals of the two programs increases significantly with the increase in atomic spacing, the most likely misaligned parameter is: A. Integral storage format\ \ \ \ B. Schwarz screening threshold\ \ \ \ C. Density fitting accuracy\ \ \ \ D. Basis set hypershrinkage parameters \\
\small Answer: B. Schwarz screening threshold \\
\midrule
\small \textbf{H5} \\
\midrule
\small \textit{\textbf{Sample 1}:} Question: Define $g(x)$ as the highest power of 5 dividing x. For $n > 0$, $S_n = \sum_{k=1}^{5^{n-1}} g(5k)$. If the largest $n < 1000$ with $S_n$ a perfect square is 499, what is X? A. X = 2\ \ \ \ B. X = 3\ \ \ \ C. X = 5\ \ \ \ D. X = 7 \\
\small Answer: C. X = 5 \\
\small \textit{\textbf{Sample 2}:} Question: When the photon energy approaches the Planck scale, the influence of quantum fluctuations on Einstein's field equations is mainly reflected in: A. Renormalization of energy-momentum tensors\ \ \ \ B. Fractional-dimensionalization of differential structures\ \ \ \ C. Topological invariance breaking of the connection term\ \ \ \ D. Supersymmetric extension of the curvature tensor \\
\small Answer: A. Renormalization of energy-momentum tensors \\
\end{longtable}

\end{document}